\documentclass[journal]{IEEEtran}

\usepackage[pdftex]{graphicx}
\usepackage[cmex10]{amsmath}
\usepackage{amssymb}
\usepackage{array}
\usepackage[caption=false,font=footnotesize]{subfig}
\usepackage{url}
\usepackage{epstopdf}
\usepackage{multirow}
\usepackage{cite}
\usepackage{soul}
\usepackage{booktabs}

\usepackage{color}

\def\act{_{\rm act}}
\def\ndofs{{n_{\rm dofs}}}
\def\task{_{\rm task}}

\begin{document}
\title{Assessing Whole-Body Operational Space Control in a Point-Foot Series Elastic Biped:\\ Balance on Split Terrain and Undirected Walking}
\date{\today}
\author{Donghyun~Kim,
   \and Ye Zhao,
   \and Gray Thomas, and
   \and Luis~Sentis
\thanks{All authors are with the Department of Mechanical Engineering, The University of Texas at Austin, Austin TX 78712-0292}
\thanks{D. Kim, Y. Zhao, G. Thomas, and L. Sentis are with the Human Centered Robotics Laboratory}
}

\markboth{IEEE Journal on XX,~Vol.~XX, No.~YY, \today}%
{Shell \MakeLowercase{\textit{et al.}}: Bare Demo of IEEEtran.cls for Journals}

\maketitle

\begin{abstract}
In this paper we present advancements in control and trajectory generation for agile behavior in bipedal robots. We demonstrate that Whole-Body Operational Space Control (WBOSC), developed a few years ago, is well suited for achieving two types of agile behaviors, namely, balancing on a high pitch split terrain and achieving undirected walking on flat terrain. The work presented here is the first implementation of WBOSC on a biped robot, and more specifically a biped robot with series elastic actuators. We present and analyze a new algorithm that dynamically balances point foot robots by choosing footstep placements. Dealing with the naturally unstable dynamics of these type of systems is a difficult problem that requires both the controller and the trajectory generation algorithm to operate quickly and efficiently.
We put forth a comprehensive development and integration effort: the design and construction of the biped system and experimental infrastructure, a customization of WBOSC for the agile behaviors, and new trajectory generation algorithms.
Using this custom built controller, we conduct, for first time, an experiment in which a biped robot balances in a high pitch split terrain, demonstrating our ability to precisely regulate internal forces using force sensing feedback techniques. 
Finally, we demonstrate the stabilizing capabilities of our online trajectory generation algorithm in the physics-based simulator and through physical experiments with a planarized locomotion setup. 
\end{abstract}


\newcommand*\mystar{{\protect \includegraphics[width=0.8em]{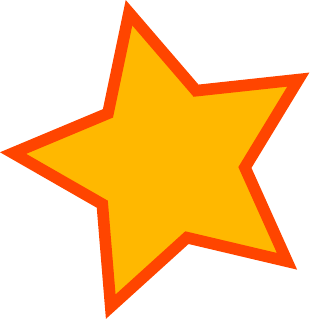}}}
\newcommand*\mystance{{\protect \includegraphics[width=0.8em]{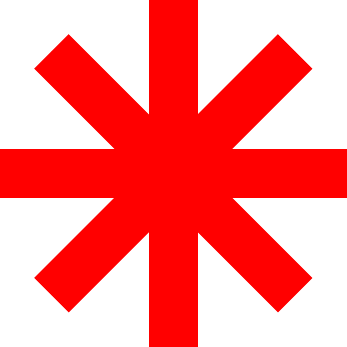}}}
\newcommand*\mystart{{\protect \includegraphics[width=0.8em]{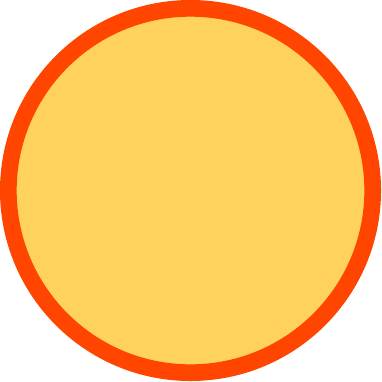}}}
\newcommand*\myfin{{\protect \includegraphics[width=0.8em]{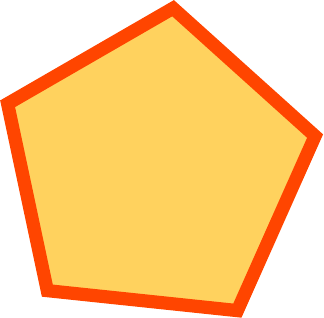}}}
\newcommand*\mynext{{\protect \includegraphics[width=0.8em]{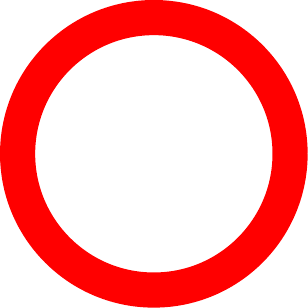}}}
\newcommand*\myimpact{{\protect \includegraphics[width=0.8em]{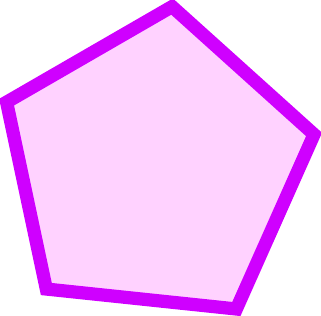}}}


\section{Introduction}

Controlling contact with the external world is critical for smooth operation of bipedal robots in cluttered environments, quickly climbing rough surfaces, and safely colliding with objects.
Addressing these goals necessitates that we understand the hardware and computational requirements of general agile behaviors and that we validate such behavior in physical systems.
The main objective of this paper is to advance agility by leveraging Whole-Body Operational Space Control \cite{Sentis:10(TRO)} (WBOSC). We focus on dual contact maneuvers in extreme terrain, and undirected walking.
We achieve this by (1) building a hardware infrastructure based on a series elastic point foot bipedal robot, (2) leveraging WBOSC to regulate internal force behavior and achieve dynamic locomotion, (3) developing a new trajectory generation algorithm for undirected walking, (4) testing dynamic locomotion with a physics based simulation, and (5) conducting various experiments on agility.


Whole-body Operational Space Control is a framework which returns the joint torques consistent with a desired set of operational space accelerations, known contact constraints, and desired internal forces.
The internal forces, during multi-contact, correspond to the linear subspace of joint torques that do not cause accelerations of the robot. The basic contact-consistent whole-body operational space control structures were laid out in \cite{sentis2007synthesis} and then extended in \cite{Sentis:10(TRO)} but only demonstrated in simulation. 
This differs from the work of \cite{hyon2007full}, a related strategy using torque controllers to optimize the distribution of reaction forces, in that WBOSC considers internal forces explicitly, and separately from the operational space acceleration goals, and places them under feedback control. This feature also separates WBOSC from the controller of \cite{Stephens-Thesis:2011}, which acknowledges the internal forces but specifies reaction forces and does not impose a feedback law on the internal component. Implementation of a whole-body controller for quadrupedal robots with optimized distribution of the reaction forces is described in \cite{hutter2013hybrid}. Hardware experiments on a series elastic actuated quadruped robot with actuated ankles using quadratic programming to minimize tangential reaction forces are shown in \cite{righetti2013optimal}. Approaching the problem from a planning perspective, \cite{todorov2011convex} explores advanced optimization methods to solve the multi-objective contact dynamics of graphical avatars. Recently this team has begun to test quasi static contact tasks in small size humanoid robots. The above is just a short list of whole-body controllers of similar type. There is an abundance of whole-body task controllers for legged  robots that we shall review further below. In view of the available whole-body controllers, the work presented here is the first to implement a whole-body operational space controller on a point foot biped robot, it is the first to show biped balancing on high pitch split terrains, and is the first to use whole-body operational space control for biped dynamic locomotion.

In essence, the main contributions of this study are: (1) creating a bipedal robot infrastructure based on whole-body operational space control, (2) incorporating sensor-based feedback controllers for internal force regulation during balancing, (3) introducing a new online trajectory generation algorithm for undirected walking, and (4) assessing the performance of whole-body operational space control for balancing on a high pitch split terrain and for undirected walking.

\section{Related Work}

\subsection{Control of Robots with Point Feet}
Point-foot biped robots similar to ours have been widely used for dynamic locomotion \cite{pratt2001virtual, westervelt2007feedback, busspreliminary, chevallereau2003rabbit, yang2009framework, hereid2014dynamic} due to their mechanical simplicity. Point-foot robots have the potential to accomplish more agile swing trajectories without the weight of an ankle. Point feet, however, are more difficult to control than robots with powered ankles since they lack the ability to apply a torque to the ground. Few have managed to walk upright without the help of a constraint mechanism, the two most notable examples being the hydraulically actuated hopper from \cite{Raibert-book:86} and the biped from \cite{busspreliminary}. This fundamental difficulty can still be meaningfully addressed in a planar case, and the vast majority of research on point foot biped robot locomotion focuses on robots constrained to a plane.

Similar to \cite{grizzle2013progress} our robot is electrically powered and has six actuated leg joints. In terms of servo rates, point foot biped systems are especially unforgiving, since it is difficult to reduce the fundamental time constant for falling over. This makes sensing systems for point foot bipeds different from those of robots with actuated ankles, such as Atlas \cite{fallon2014architecture}, Valkyrie \cite{paine2014actuator}, and Sarcos \cite{Stephens:2010fj} in that they must be focused on fast maneuvers. To accommodate their fast dynamics, special care must be taken to ensure that the control systems for these robots operate at high frequency and that motion is detected at high speed. Relative to the infrastructure of other point foot robots such as \cite{sreenath2011compliant} and \cite{ hereid2014dynamic}, our setup includes an additional overhead motion capture system, which communicates with the control computer to allow absolute position feedback. We do not have sensors on the planarizer's joints or slider.

\subsection{SEA Control}

Series Elastic Actuators (SEAs) are designed with an elastic element and provide two important benefits over their rigid, or directly connected, counterparts: they offer improved force control accuracy \cite{pratt1995series,vallery2007passive}, and a lower output inertia. This high fidelity torque tracking is critical to contact scenarios where the internal multi-contact forces must be maintained within a certain range, as is the case when a robot balances on steep terrain. Some robots, for example the robots of \cite{sreenath2011compliant} and \cite{ grizzle2013progress}, use SEAs only for energy storage, and do not take advantage of their force control capabilities. Hume, on the other hand uses SEAs primarily for their force control and inertial benefits to handle collisions.
Yet the inclusion of SEAs comes at a cost to performance. Controllers that have been designed to approximate the dynamics of a perfect position source, a perfect torque source \cite{pratt1995series, vallery2007passive,painedesign2014}, or a second order mass-spring-damper system all face the same problem in their final closed loop system: the dynamics of the reference plant are impossible to achieve at the highest frequencies. 

The use of SEAs for biped locomotion was pioneered in \cite{pratt1997stiffness}, which suggests a straightforward PD torque control strategy with some model-based feedforward terms. However, the work does not compensate for static errors. The same torque controller was used in the SEA actuated robot \cite{pratt2012capturability} to study push recovery. However, such an application does not require accurate force tracking so much as position control, and thus it is not possible to asses the torque performance of these actuators. Trajectory following accuracy did appear to be limited by the SEA compliance; In the author's own words they state: ``Due to the use of SEAs with very compliant springs, we have had difficulty to quickly and accurately swing the leg,'' A sentiment echoed by our own observations. Recent studies describe potential solutions for this type of problem with SEA actuators \cite{ye2014sea, spong2006robot, vallery2007passive, mosadeghzad2012comparison}, with passivity based controllers emerging as a solid, if conservative, approach. \cite{vallery2007passive} adds an inner motor velocity loop and incorporates integral torque action to the controller. More recently, \cite{mosadeghzad2012comparison} compared the stability and performance of various active impedance control approaches based on cascaded SEA controllers. 

\subsection{Whole-Body Controller Design}

Most existing experimental approaches for whole-body control are based on optimization methods which were pioneered by \cite{hyon2007full}. \cite{Stephens-Thesis:2011} represents the first implementation of full dynamic based task controllers with contact constraints on a humanoid robot. The work focuses on push recovery and basic walking. In \cite{Herzog:2014} the implementation of hierarchical inverse dynamics algorithms using quadratic program solvers is presented on a Sarcos biped robot. Experiments include balancing while withstanding external forces, balancing on a moving platform, and balancing on a single foot. In \cite{Bertrand-Pratt:2014} a torque-based whole-body controller is presented for controlling the Atlas and Valkyrie humanoid robots. A quadratic program solver is used to minimize momentum rate objective, contact force regulation and task acceleration regularization. In \cite{saab-tro-13} whole-body control with inequality constraints via inverse dynamics and a quadratic program solver is proposed on the humanoid robot HRP-2. However, the algorithm is used offline to generate trajectories that are then tracked by the robot. In \cite{Henze-Ott-Roa:2014} a torque-based whole-body controller focused on optimization of multicontact wrenches over a period of time corresponding to center of mass (COM) movement is presented. The approach is, for now, presented in the context of balancing and is only shown in simulation. All of these works aim only to control humanoids with actuated ankles, and thus do not consider the under-actuation situations endemic to dynamic point-foot walking. Separately, in \cite{hutter2014quadrupedal} optimization methods based on inverse dynamics projected on the contact null space are used to control the gait of a quadrupedal robot. 
In contrast to these works, ours is the first to implemented an operational space inverse dynamic algorithm in an underactuated bipedal point foot robot. It does not attempt to control the center of mass during locomotion, relying solely on estimation via prismatic inverted pendulum dynamics. Additionally, the previous works have not implemented sensor-based internal force control nor have they attempted to balance the robots on such extreme surfaces. 

In particular, between those controllers that perform task space inverse dynamics, there are controllers which calculate internal forces as a byproduct of another optimization and there are those that require the specification of the internal forces of contact. Methods which deal with contact forces as generally interested in balance \cite{stephens2010dynamic, lee2012momentum, hyon2007full} rather than in controlled interaction with the environment. For instance maintaining a stance between highly sloped surfaces. In their work the main objective is to maintain balance while keeping the reaction forces within friction cones so contact is maintained. \cite{hutter2014quadrupedal} used the reaction force method to keep a quadruped on a surface with a approximately $40^\circ$ inclinations. However, until now, no group has approached to problem of accurately controlling these internal forces through feedback control. Previous methods have fed the desired reaction forces into their inverse dynamics routine as a feed-forward term, and the error between the actual internal force and the achieved internal force was not used to lower this error through joint torque. Such feedback would potentially require significant modification to optimization based method of determining the feed-forward reaction force. 

Whole-Body Operational Space Control \cite{sentis2007synthesis}, is centered around the idea of achieving operational space impedance control and controlling the internal forces between supporting contacts. In particular, internal forces are chosen such that they are decoupled from the motion of the robot and can be directly specified by a high level planner. By using feedback control on these internal forces we can achieve higher precision force tracking which is less sensitive to modeling errors than it would if we used feed-forward internal forces. 

\subsection{Locomotion}

A hybrid dynamical systems problem, point-foot locomotion in the plane is difficult because single support motion is under-actuated and thus the system can only be stabilized when the discrete effects are taken into account.\footnote{Provided we are willing to neglect the extra controllability afforded by the Coriolis and gravity terms.}
One of the most successful approaches to this problem comes from \cite{sreenath2011compliant}, wherein the feedback whole body controller constrains the dynamics to match a model which is predicted by a simulation to be stable in the discrete sense. Another successful approach based on hybrid zero dynamics is the line of work by \cite{Ames-Amber2:2014} which utilizes human inspired trajectories to generate stable periodic locomotion. These formulations are designed to achieve periodic motions, as is also the case for other works based on Poincare maps \cite{hereid2014dynamic}. 

Algorithms such as the capture point \cite{pratt2012capturability} and phase space planning approaches \cite{zhao2012three} can be used to stabilize the pendulum model by modifying footstep locations. Phase space planning extends the linear pendulum model to consider the case where the center of mass height is a function of the horizontal position, the prismatic inverted pendulum model \cite{donghyun2014icarcv}. Thus one objective of this paper is to assess the performance of our whole body controller in the task of restricting the dynamics to match those of a prismatic inverted pendulum model. In this last conference paper we presented an earlier version of the trajectory generation algorithm considered in this paper. However, no control of internal forces or experiments of any type were conducted. 
As stated in \cite{Grizzle2014Automatica} there is still little work on aperiodic walking. Our trajectory generation algorithm achieves aperiodic gaits by exploiting a simple time to velocity reversal rule. It is in some ways close to the algorithm by~\cite{Pratt-Dilworth-Pratt:97} but designed for balancing in 3D instead of walking in the 2D plane.


\begin{figure}
\includegraphics[width = 8cm]{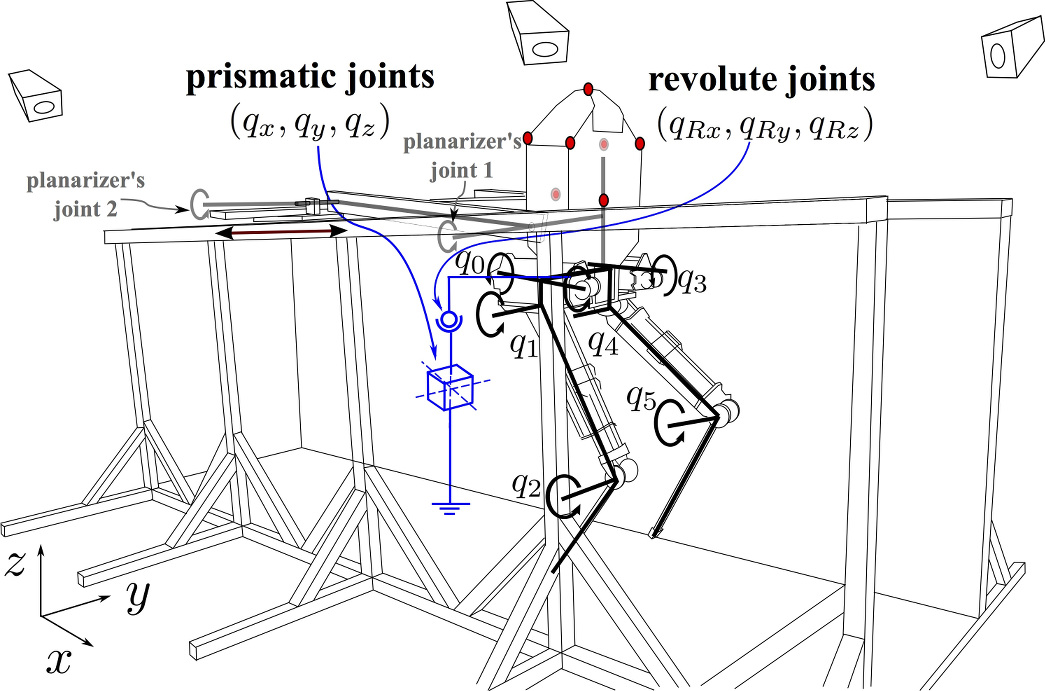}
\caption{\textbf{Hume Robot Kinematics,} as when attached to the planarizing linkage. Blue schematics describe the floating base joints, while black describes the kinematics of the six leg joints and three planarizer joints. The planarizer kinematics are not included in the generalized joint vector, since they are not part of the robot's model. The locations of the LED tracking markers identified by the PhaseSpace Impulse Motion Capture system are shown as red dots.}
\label{fig:infra}
\end{figure}

\begin{figure*}
\centering
\includegraphics[width = 0.9\linewidth]{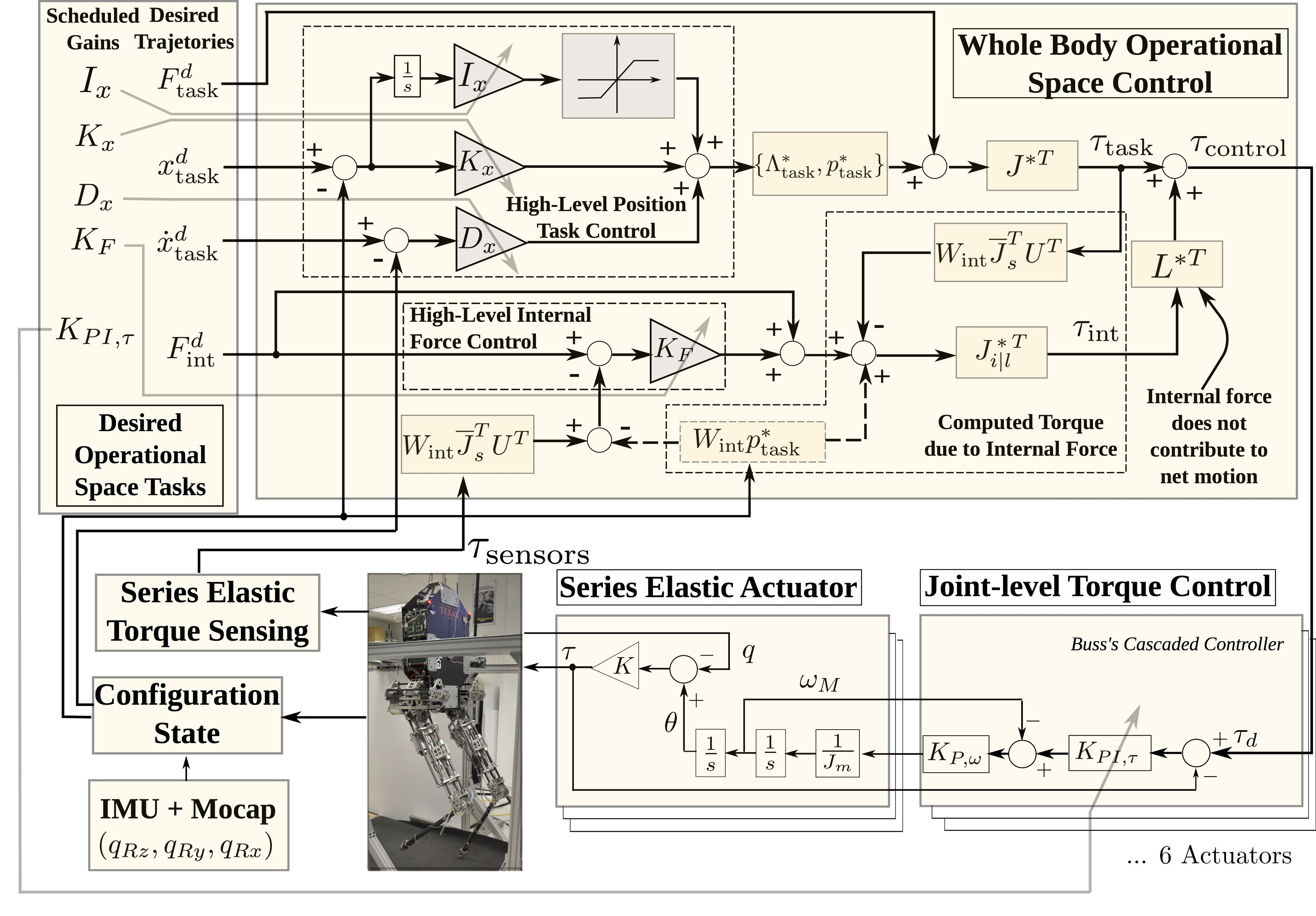}
\caption{\textbf{Overall Control Diagram.} This figure illustrates the whole body operational space control process (WBOSC) and the joint-level torque controllers. One of our main contributions comes from the feedback control of internal forces. Note that the gains for the controllers are treated as additional input parameters to represent the gain scheduling we perform in order to achieve the best possible performance with each task.}
\label{fig:control}
\end{figure*}

\section{System Characterization}
\subsection{Hardware Setup}
Our robot is a teen-size humanoid robot measuring 1.5 meters in height and 20 kg in weight. The leg kinematics resemble simplified human kinematics and contain an adduction/abduction hip joint followed by a flexion/extension hip joint followed by flexion/extension knee joint as shown in Fig.~\ref{fig:infra}. The lack of ankle joints allows the shank to be essentially just a lightweight carbon fiber tube. At the tip of the shank we have incorporated contact sensors which are essentially limit switches. The series elastic actuators on all six joints are based on a sliding spring carriage connected to the output by steel ropes. The deformation in the springs are directly measured within the carriage assembly. The concept, kinematics, and specifications of the robot were proposed by our team at UT Austin, and the design was executed in collaboration with Meka Robotics and manufactured by that company. For fall safety the robot is attached to a trolley system with a block and tackle system which allows easy lifting and locking at a height. In addition, the robot has a removable planarizer mechanism which constrains the motion of the robot to the saggital plane. The pitch of the robot remains unconstrained, as the planarizer connects to the robot through a set of bearings. Ultimately, pitch, forward motion, and vertical motion are allowed, while lateral motion, roll, and yaw are prohibited.

From an electrical point of view the robot is controlled with distributed digital signal processors, connected by an EtherCat network to a centralized PC running a real-time RTAI Linux kernel. This communication system introduces a 1ms delay from the linux machine to the actuator DSPs and back. Each DSP controls a single actuator, and they do not communicate directly with each other. Power is delivered through a tether from an external source.

A Microstrain 3DM-GX3-25-OEM inertial measurement unit (IMU) on the robot's torso measures angular velocity and linear acceleration, which is used in the state estimator. Additionally, the robot has a full overhead PhaseSpace Impulse motion capture system which gives it global coordinate information about seven uniquely identifiable LED tracking devices mounted rigidly to the torso. The PhaseSpace system produces a data stream at 480Hz and communicates to the Linux Control PC via a custom UDP protocol. There is an approximate 15ms delay in the data for feedback purposes. It accomplishes this using a system of sixteen high speed cameras mounted on the ceiling above the robot, and a proprietary software package to fuse their readings into a single estimate for the three dimensional position of each marker. On each update, the system reports the location of as many of the uniquely identifiable LED markers as it can see. 

\subsection{End-to-End Controller Architecture}
The feedback control system is split into six joint level controllers and a centralized high level controller (see Fig.~\ref{fig:control}). The purpose of the joint level controllers is to achieve good torque tracking given the series elastic actuators. This type of control architecture falls into the category of a distributed control system which allows the joint controllers to focus on high speed actuator dynamics while the centralized controller does not need to deal with this nuance. Yet the feedback at the high level is necessary in order to create the coupling between joints implied by operational space impedance tasks as well as regulating the internal forces between multicontact supports. 

\subsection{Series Elastic Actuators}
\label{sec:gains}
The robot came with MEKA's pre-loaded joint controllers based on the passivity torque controller described in \cite{vallery2007passive} (shown on the lower right corner of Fig.~\ref{fig:control}). We kept this controller's structure while changing the gains of the feedback controller to enhance the performance of the high level controller, and this ultimately entailed reducing the low level torque gains. In order to tune the torque gains we leveraged our findings in \cite{ye2014sea}. In this study we describe a trade-off between torque gains and position gains in a distributed control architecture. Specifically, we explore the observation that raising the torque controller's proportional gain limits the maximum stable position gains and vice versa. To respond to this observation we implemented a gain scheduling strategy: in the joints of the stance leg we lowered the torque gains so we could raise the proportional gain and reduce error. In the joints of the swing leg we raised the torque gains to produce less friction dominated behavior.

\subsection{Whole Body Operational Space Control}

Whole-Body Operational Space Control \cite{Sentis:10(TRO)} is a feedback control strategy based on Operational Space Control \cite{Khatib87}, which extends it to floating base robots in contact with the environment. It allows the user to specify multiple task objectives and their impedance in operational space. It additionally subdivides the torques applied to the robot into orthogonal spaces which affect either the motion of the robot or the internal forces which do not. When the user specifies these internal forces our implementation governs them using feedback. Whole-Body Operational Space Control is explained in Appendix \ref{sec:appendice}.

At the implementation level, WBOSC worked -- provided that our latencies were sufficiently small. Achieving a $1\ \rm ms$ latency required some significant software work. We re-wrote the firmware provided by our robot manufacturer in order to ensure our controller operated within a real-time thread, and to incorporate it into a hierarchical chain structure which ensures minimum latency for stacked control systems. We also reduced the basic computational cost of our WBOSC algorithm by bypassing recursive dynamics software and instead using a closed form expression to calculate the mass matrix. In order to reduce the tracking error, we added an integral term to all position tasks, which helps alleviate the friction difficulties involved in lowering the torque gains at the DSP level. This also reduces error due to inaccuracy in the gravity estimation term and other steady state errors in our dynamics model. 

When operating with the robot within a planarizing linkage mechanism, as is the case for the experimental section of this paper, we still model the robot as having a floating base. We then incorporate the additional inertia of this planarizer as a lumped mass inside the floating base.
Thus, in the equations in Appendix \ref{sec:appendice}, the generalized coordinate variable $q$ corresponds to the six degrees of floating base kinematics plus six degrees of freedom corresponding to abduction, hip, and knee kinematics for the two legs as shown in Fig.~\ref{fig:control}. Our matrix $U$ corresponds to floating base kinematics.
We also added terms to the mass matrix representing the rotor inertia of the motors, which showed a slight improvement in performance. 
Finally, we tuned the task gains experimentally using heuristics.

\subsection{State Estimation and Sensor Fusion}
\label{sec:sensing}
The controller needs an estimate of the body orientation every millisecond, yet the motion capture system updates at 480 Hz, occasionally fails to track a subset of the markers, and has a non-trivial latency of 15 ms. Given that the on board IMU accurately reports the rotational velocity of the torso with respect to an inertial reference frame, we integrate forwards in time to maintain an estimate of the orientation while waiting for an update from the overhead positioning system. When such an update arrives, we acknowledge the feedback latency of the sensing system process, and generate an innovation measurement based on not the most current value of the orientation estimate, but the estimate from 15 controller update periods into the past \-- i.e. the ratio between the 15ms latency of the motion capture system and the 1ms servo rate.

Calling this time $k= t-15$, we can setup a least squares problem which minimizes the distance between measured LED position 
$\hat y^k_i \in \mathbb{R}^{3}$ and predicted LED position 
$\tilde y^k_i \in \mathbb{R}^{3}$ for $i=1,\dots, n$, where $n$ is typically 7, but decreases when LEDs are blocked or otherwise non-visible. Our model predicts LED locations based on an affine transformation of a default pattern
$\tilde y^k_i = x^k + A^k z^k_i$ where
$T^k=\{x^k\in \mathbb{R}^{3},A^k\in \mathbb{R}^{3\times3}\}$ is the affine transform at time $k$ and the default pattern, $z_i \in \mathbb{R}^{3}$, is essentially just the measurement from some default position, but is shifted such that the center of the coordinate system is the geometric center of the points. That is the first moment is zero for the pattern:  $\Sigma_{i=1}^7 e_j z_i=0$ for $j=0,1,2$. This affine transform includes both a linear bias term which represents translation of the geometric center of the LEDs, and a matrix term which represents rotation as well as non-physical skewing and scaling of the pattern. The pattern is specifically designed such that the origin of the pattern frame is the geometric center of the points. As this model is linear in the affine parameters we can solve for the case where all the LEDs are visible as follows
\begin{gather} \label{eqn:y_def}
	\theta^k=
		\begin{pmatrix} 
			x^k\\
			\mathrm{vec}(A^k)
		\end{pmatrix} 
,\qquad
	\tilde {\mathbf{y}}^k = 
		\begin{pmatrix} 
			\tilde y^k_1\\
			\vdots \\
			\tilde y^k_7
		\end{pmatrix} =
	R \theta^k,
\\[3mm]\label{eqn:R_def}
	R = 
		\begin{pmatrix}
			I_{3\times3} & e_0 z_1^\intercal  & e_1 z_1^\intercal & e_2 z_1^\intercal\\
			I_{3\times3} & e_0 z_2^\intercal & e_1 z_2^\intercal & e_2 z_2^\intercal\\
			\vdots &\vdots&\vdots&\vdots \\
			I_{3\times3} & e_0 z_7^\intercal & e_1 z_7^\intercal & e_2 z_7^\intercal \\
		\end{pmatrix}, 
\\[3mm]\label{eqn:theta_simp}
	\theta_{simp}^k \triangleq (R^\intercal R)^{-1}R^\intercal \hat{\mathbf{y}}^k, 
\end{gather}
where \eqref{eqn:y_def} describes affine transforms in vector form, and demonstrates the linearity of prediction, and \eqref{eqn:theta_simp} defines the affine transform which best describes the observed LED vector $\hat{\mathbf{y}}^k$ as a transform of the pattern. However, we have opted to lowpass this sensor data by adding twelve extra rows of regularization terms and a diagonal weighting matrix, 
\begin{equation}\label{eqn:weighting}
	W=
		\begin{pmatrix}
			I_{3n\times3n} & 0 & 0\\
			0 & \lambda_1 I_{3\times3}&0\\
			0 & 0 & \lambda_2 I_{9\times9}
		\end{pmatrix} ,
\end{equation}
to the least squares equation. In addition, we cannot always assume that all LEDs are visible and we must define a knockout matrix $K_o$ which selects the LEDs which the system successfully located:
\begin{equation}
K_{o} \in \mathbb{R}^{n\times 7}= 
	\begin{pmatrix}
		e_0^\intercal \text{ if LED 1 was found}\\
		e_1^\intercal \text{ if LED 2 was found}\\
		\vdots\\
		e_6^\intercal \text{ if LED 7 was found}
	\end{pmatrix}.
\end{equation}
Where $(K_{o} \otimes I_{3\times3})$ selects only equations relating to observed LEDs from the original regressor, 
\begin{equation}\label{eqn:R_r}
	R_r \triangleq
		\begin{pmatrix}
			(K_{o} \otimes I_{3\times3}) R \\ I_{12\times12}
		\end{pmatrix}.
\end{equation}
This results in a new estimate of the affine transform, 
\begin{equation} \label{eqn:reg_lstsq}
	\theta_r^k \triangleq (R_r^\intercal W R_r)^{-1}
	R_r^\intercal W
		\begin{pmatrix}
			(K_{o} \otimes I_{3\times3}) \hat{\mathbf{y}}^k \\
			\tilde \theta({k|k-p})
		\end{pmatrix},
\end{equation}
where the integration of the IMU orientation rate data, 
\begin{equation} \label{eqn:imu_int}
	\tilde \theta({b|a})=\theta_r^{a}+\sum_{t=a}^b
		\begin{pmatrix}
			0_3 \\
			 \mathrm{vec}(\hat{\omega}_{\mathrm{IMU}}^t\times) 
		\end{pmatrix} \Delta t,
\end{equation}
is the basis for the regularization term setpoints, as shown in \eqref{eqn:reg_lstsq}. 

The regularization adds another term to the objective function, specifically the squared deviation between each element of the affine transform rotation matrix and our estimate of this matrix given the orientation estimate at time k. Regularization is also applied to the vector component of the affine transform, but the weight $\lambda_1$ is so small as to be neglected. The weight $\lambda_2$, on the other hand, represents a significant initial covariance which specifies the tradeoff between old knowledge and new knowledge. The weights were chosen such that the discrete time half life of an error in the orientation estimate is one update if all seven LEDs are visible. 

Solving this least squares problem returns a transform which is potentially skewed and scaled as well as rotated, and thus is not a valid rigid body transform. By finding the closest quaternion to this rotation matrix we constrain the result to valid transforms, using the method of \cite{bar2000new}. Finally, with the orientation estimate at time k being the closest quaternion to the affine transform, we can integrate the IMU data from times k to t to estimate the orientation at time t. The integration of IMU data continues incrementally while the algorithm waits for the next piece of overhead camera data to arrive, and then it repeats the more complex fusion algorithm. Since the algorithm performed adequately the first time it was used, no attempt was made to precisely tune the delay estimate and filter constants.

\subsection{Contact Transitions}
\label{sec:transitions}

In order to reduce the high speed behavior associated with a sudden change in joint torques, we have implemented a strategy which acts to smooth out the torque commands when the robot transitions between single and dual support.
The sudden change of torque commands is due to the instantaneous switch between constraint sets within WBOSC, and our method to smooth these torque commands effectively bridges the difference between single contact and dual contact.
To make WBOSC with a single contact constraint produce the result it would with a dual contact constraint we simply add a desired reaction force to the swing foot -- the same reaction force that would be expected of this foot in the dual contact case. Then, to transition, we decrease this desired reaction force linearly with time in the case of foot lifting, or linearly ramp it up from zero in the case of foot landing. This requires that we know this reaction force beforehand, so we must always run the controller with dual contact constraints before the single contact version. When lifting the foot we can use the previous value of the reaction force, but when landing we run the controller in dual contact once at the beginning of the foot landing transition phase for the sole purpose of acquiring this reaction force.

To implement this desired external force in single contact WBOSC we add the term $f_{ext}$ to Equation~\eqref{eq:task-space-dynamics},
\begin{equation}
F\task = \Lambda\task^* u\task + \mu\task^* + p\task^* + f_{ext}. 
\end{equation}
This force $f_{ext}\triangleq w f_{ext, \rm \;dual}$ can be extracted from the dual contact WBOSC after the output torque is calculated based on Equation~\eqref{eq:constraintforces}
\begin{equation}
f_{ext, \rm \;dual} = S_{\rm swing} \left(\overline J_s^T \left[U^T \tau_{\rm control} -b -g\right] + \Lambda_s \dot J_s \dot q \right),
\end{equation}
where $S_{\rm swing}\in \mathbb{R}^{3\times6}$ selects the constraint forces of the swing leg from those of both foot contact constraints and $w\in[0,1]$ represents the linear ramp.

\section{Feedback-Based Internal Force Control}
\label{sec:internal_force}
\begin{figure}
\centering
\includegraphics[width = \columnwidth] {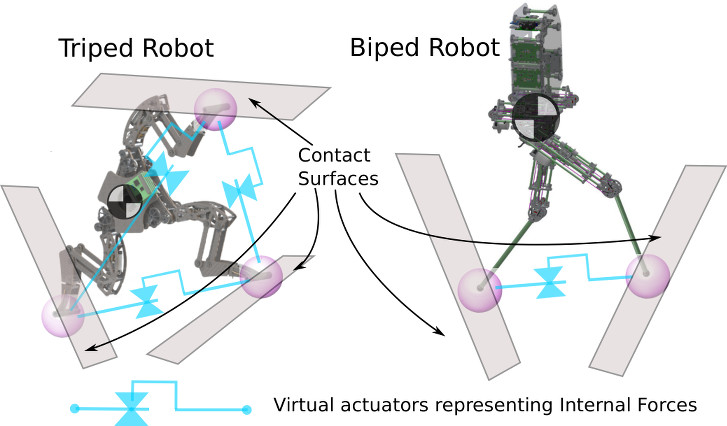}
\caption{{\bf Illustration of Internal Forces for Various Robots.} Internal forces in point foot robots correspond to tensions or compressions between pairs of supporting contacts.}
\label{fig:internal_forces}
\end{figure}

Internal force behavior corresponds to actuator forces that produce no net effect on the robot's motion. A such, internal forces correspond to mutually canceling forces and moments between pairs or groups of contact branches, i.e. tensions, compressions and reaction moments. For instance, a triped point foot robot has three internal force dimensions while a biped point foot robot has a single internal force dimension as shown in Figure~\ref{fig:internal_forces}. 

In this context, building a biped robot with excellent torque sensing has two advantages: (1) its ability to use low level torque control to overcome actuator friction and achieve greater control bandwidth. In turn, our rigid body assumptions to model internal forces are less affected by low level actuator dynamics; and (2) torque sensors on the robot's joints permit the implementation of sensor-based internal force feedback for accurate tracking.  

What is interesting about taking a model-based approach is that internal forces are fully controllable by definition as they are orthogonal to the robot's motion. As such, both the robot's movement and its internal forces can be simultaneously controlled to feasible values. Moreover, in many types of contact poses, internal forces are easily identifiable using some physical intuition. For instance, in the triped pose of Figure~\ref{fig:internal_forces} the three feet can generate three virtual tensions between the points of contact. The physics of tension forces were analyzed in greater detail using a virtual linkage model in~\cite{Sentis:10(TRO)}.

Internal forces are part of the core functionality of Whole-Body Operational Space Control. In the Appendix section, we describe the model-based control structures enabling direct control of internal forces. 
In particular, the basic torque structure derived in Equation~(\ref{fig:int-for}) is written here again for readability,
\begin{equation}
\Gamma_{\rm int} \, = \; J^{\,*\,T}_{i|l} \Big(F_{\rm int, ref} -  F_{{\rm int},\{t\}} + \, \mu_i + \, p_i\Big),
\end{equation}
where $F_{\rm int, ref}$ is the vector of desired internal forces and $F_{{\rm int},\{t\}}$ corresponds to the mapping of task torques into the internal force manifold.
The above equation is based on the assumption that commanded torques and actual torques are identical, and that the kinematic and dynamic models are exact. Because these premises are never true, to achieve best results on force regulation or tracking, we propose to employ sensor-based feedback control on the internal forces.
To our knowledge, this is the first time that sensor-based feedback control of internal forces is proposed and achieved in a real robot. 

\begin{figure*}
\centering
\includegraphics[width = 1.9\columnwidth]{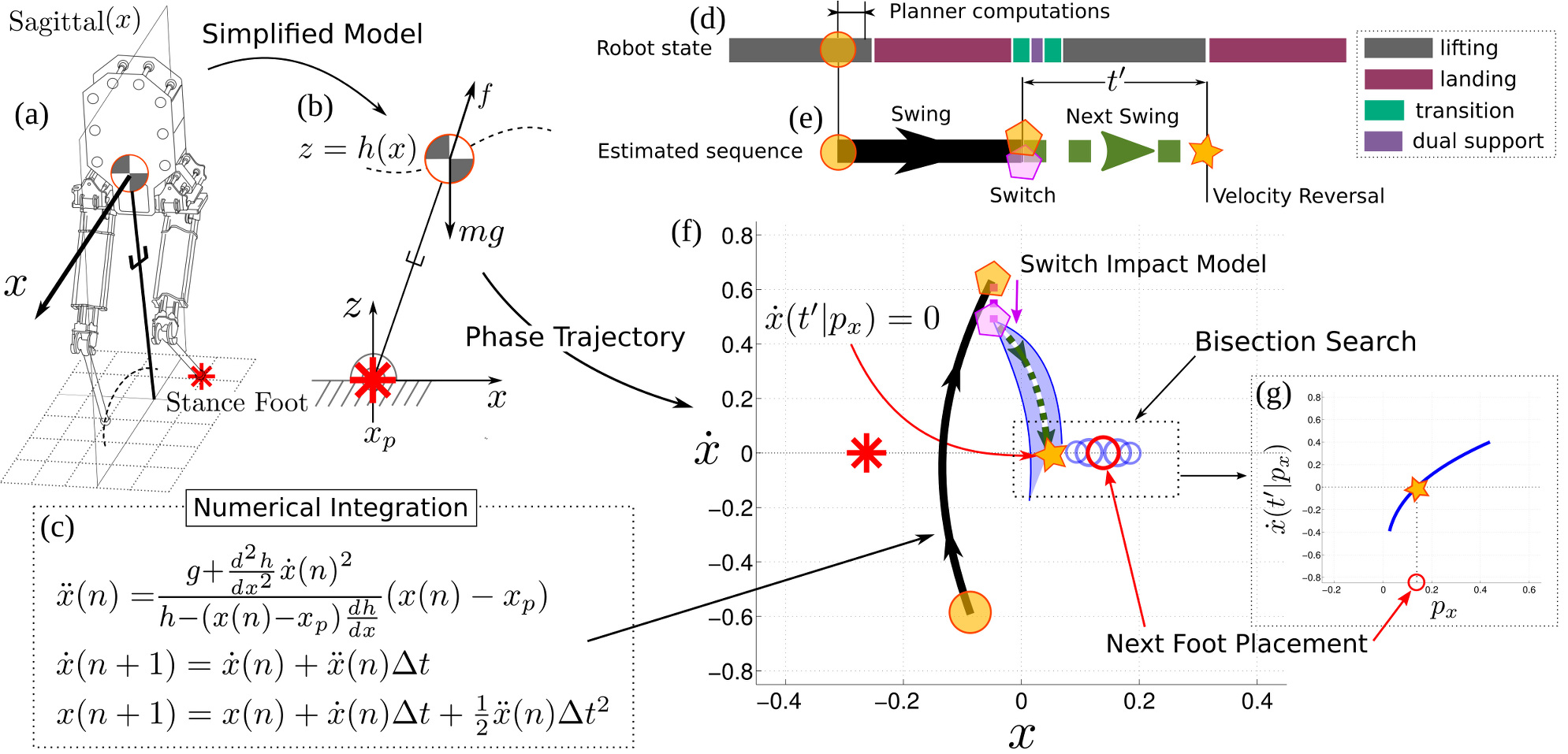}%
\caption{{\bf Constant Time to Velocity Reversal Algorithm.} As shown in (a), we approximate the dynamics of the robot with the prismatic inverted pendulum, shown in (b). This model predicts the dynamics of the horizontal center of mass position $x$ given the stance foot location $x_p=\ $\mystance\ and the height surface $z=h(x)$. This can be integrated forward in time via the numerical integration procedure shown in (c). When the planner starts operating it records the initial state \mystart \ and integrates this state forward to determine the switching state \myfin . As shown in timelines (d) and (e), the ``Estimated sequence'' of the planner has an analogue in the ``Robot states'' of the state machine. In particular, the switching state \myfin\ roughly corresponds to the dual support phase of the walking state machine. When planning with the physical robot, the planner computes a post-impact state \myimpact\ by applying a velocity adjustment to the switching state. This empirical measure compensates for what appears to be a nearly constant decrease in velocity at every impact. This new state \myimpact\  represents the planner's guess at the time, $x$, and $\dot x$ values immediately after the switch. The goal of the planner is ultimately to stabilize the robot, but this is implemented by choosing the next footstep such that velocity reverses $t'$ seconds after the foot switch every step. For sufficiently smooth height surfaces, the relationship between the next footstep location $p_x$ and the velocity $\dot x$ is monotonic, as shown in (g). We use the standard bisection algorithm to identify the next foot placement \mynext \ which results in a velocity reversal state \mystar \  at time $t'$, as shown in (f).
}%
\label{fig:phase_explain}
\end{figure*}

Because internal forces are fully controllable, we incorporate a simple proportional controller on the internal forces into Equation~(\ref{fig:int-for}),  
\begin{multline}\label{eq:internal-force-feedback}
\Gamma_{\rm int} \, = \; J^{\,*\,T}_{i|l} \Big(F_{\rm int, ref} -  F_{{\rm int},\{t\}} + \, \mu_i + \, p_i \\+ K_F \left(F_{\rm int,ref} - F_{\rm int, act}\right) \Big),
\end{multline}
where $K_F$ is a proportional control gain and $F_{\rm int, act}$ are the actual sensor-based internal forces. In order to obtain these sensor-based forces, we use the torque sensors on the series elastic actuators to find the co-states of constraint as per Equation~\eqref{eq:constraintforces} and apply a projections $W_{\rm int}$ to find internal forces,
\begin{equation}\label{eq:internal_act}
F_{\rm int, act} \triangleq W_{\rm int} \,  \left[ \overline{J}_s^T (U^T \tau_{\rm sensor} - b - g)+\Lambda_s \dot J \dot q\right],
\end{equation}
where $\tau_{\rm sensors}$ corresponds to the vector of torques sensed by the spring element in each series elastic actuator (see Figure~\ref{fig:control}).

Although this internal force mapping above is distinguished from previous work, due to its sensor-based force feedback control, this mapping is valid due to the physical fact of robot redundancy in the multi-contact case. The induced contact closed loop causes the number of controlled tasks to be smaller than that of actuated joints. Correspondingly, additional DOFs are available to be controlled for more tasks, such as internal forces in Equation~(\ref{eq:internal_act}). This mapped internal force is consistent with contact constraints and cancellation of accelerations on the robot's base or on the actuated joints \cite{Sentis:10(TRO)}. More details can be found in the Appendix section.


To calculate internal forces for our bipedal robot, Hume, we need to define the mapping given in Equation~(\ref{eq:wbosc:mapping_internal_reaction}) in Appendix~\ref{sec:appendice}, where $W_{\rm int}$ is the matrix representing the map from reaction forces to internal forces. In our case, Hume controls the internal forces between the two feet during dual contact phases. In dual support, the reaction forces are $(f_{Rx},\ f_{Ry}, \ f_{Rz}, \ f_{Lx}, \ f_{Ly}, \ f_{Lz})^T$, where $R$, $L$ mean a right and left foot, respectively. According to~\cite{Sentis:10(TRO)}, $W_{\rm int}$ consists of a selection matrix of tensions, $S_t$, a rotation matrix from global frame to the direction parallel to the line between two contact points, $R_t$, and a differential operator matrix, $\Delta _t$, i.e.
\begin{equation}
W_{\rm int} = S_t \, R_t \, \Delta_t,
\end{equation}
with
\begin{align}
& S_t = \begin{pmatrix}
1 & 0 & 0 \\
\end{pmatrix}, \\[2mm]
& R_t = 
\begin{pmatrix}
\hat{x}^T \\[1mm]
\hat{y} ^T\\[1mm]
\hat{z} ^T \\
\end{pmatrix}, \quad 
\begin{cases}
\hat{x} = \frac{P_R - P_L}{||P_R - P_L||} \\[1.5mm]
\hat{y} = ( -\hat{x}(2) , \hat{x}(1), 0)^T \\[1.5mm]
\hat{z} = \hat{x} \times \hat{y}
\end{cases},\\[2mm]
&\Delta_t =
\begin{pmatrix}
I_{3\times3} & -I_{3\times3}
\end{pmatrix}.
\end{align}
where $P_R$ and $P_L$ are the position of the right and left feet, respectively.

In order to compute desired internal forces, we suggest either to use heuristics as we do in this paper, or the use of the virtual linkage model and the multicontact/grasp matrix, which were proposed in~\cite{Sentis:10(TRO)}. Those models allow to relate the center of mass and the internal force behavior to the reaction forces. A study on the usage of these models is currently beyond the intended scope of this paper.
Compared to other methods based on optimizing reaction forces, our method uses sensor-based feedback control on the internal forces to regulate or track a desired tension reference with good accuracy. 

\section{Trajectory Generation}
\label{sec:trajectory}
In order to stay upright our robot must constantly re-position its feet, and a mechanism of considerable complexity is required to decide the upcoming foot position for the swing foot. We have developed a method for finding this position which we refer to as a phase space constant time to velocity reversal planner. In every step, when the lifting phase reaches 80\% completion, the planner runs in an online fashion and returns the next footstep location before the lifting phase ends and the landing phase begins. The operational space set-point trajectory for the swing foot is then defined parametrically based on this desired landing position, with the trajectory ending once ground contact is sensed. If the planned step is outside the mechanical limits of the robot the step saturates to the closest reachable step.

The method we use to choose this footstep location operates on a one dimensional model appropriate for planar walking, but, by choosing the $x$ and $y$ components of the footstep location as solutions to the forward and lateral planar walking problems, it is extended to 3D walking. We will present first the planar algorithm used in the experimental section going on to explain our approach to 3D walking. 

\subsubsection{1D Velocity Reversal}
Our planner attempts to stabilize the robot by causing the center of mass to reverse direction every step. Central to this undertaking is the exploitation of a simplified model of the robot's zero-dynamics given specific operational space tasks: the prismatic inverted pendulum model, or PIPM. The PIPM considers a point mass constrained to an arbitrary continuous height surface by a constraint wrench which evaluates to a pure force at both the foot point and the center of mass -- that is the model assumes a point foot produces the reaction force and that the reaction force points towards the center of mass. 
The PIPM can be expressed as the differential equation
\begin{equation}
\ddot{x} = \frac{g + \ddot{z}}{z}(x -x_p ).
\label{eq:pipm_ori}
\end{equation}
Accounting for z being a function of x, the height surface, 
\begin{align}
&\frac{dz}{dt} = \frac{dz}{dx} \frac{dx}{dt},\\[2mm]
&\frac{d^2 z}{d t^2} = \frac{d}{dt}\Big(\frac{dz}{dx}\Big)\frac{dx}{dt} + \frac{dz}{dx}\frac{d}{dt}\Big( \frac{dx}{dt} \Big), \\[2mm]
&\ddot{z}  = \frac{d^2 z}{d x^2} \dot{x}^2 + \frac{dz}{dx}\ddot{x}.
\label{eq:ddotz}
\end{align}
By plugging Equation~(\ref{eq:ddotz}) into Equation~(\ref{eq:pipm_ori}), we obtain,
\begin{align}
&\ddot{x} = \frac{g + \frac{d^2 z}{dx^2}\dot{x}^2 + \frac{dz}{dx}\ddot{x}}{z}(x-x_p),\\[2mm]
&z \ddot{x} = \Big(g + \frac{d^2 z}{dx^2}\dot{x}^2 \Big)(x -x_p ) + (x - x_p ) \frac{dz}{dx} \ddot{x}, \\[2mm]
&\ddot{x} = \frac{g + \frac{d^2 z}{d x^2} \dot{x}^2}{z - (x - x_p)\frac{dz}{dx}}(x - x_p)
\label{eq:pipm}
\end{align}
which now lacks any $\ddot z$ term, and can be used in Fig.~\ref{fig:phase_explain}c to integrate forward in time.

This model is a close approximate to the zero dynamics of our robot when a specific set of WBOSC tasks are accurately maintained. These tasks are: (1) a center of mass height task, (2) a constant body link attitude, and (3) any sufficiently gentle Cartesian trajectory task for swing foot.
\begin{figure*}
\centering
\includegraphics[width = 0.95 \textwidth] {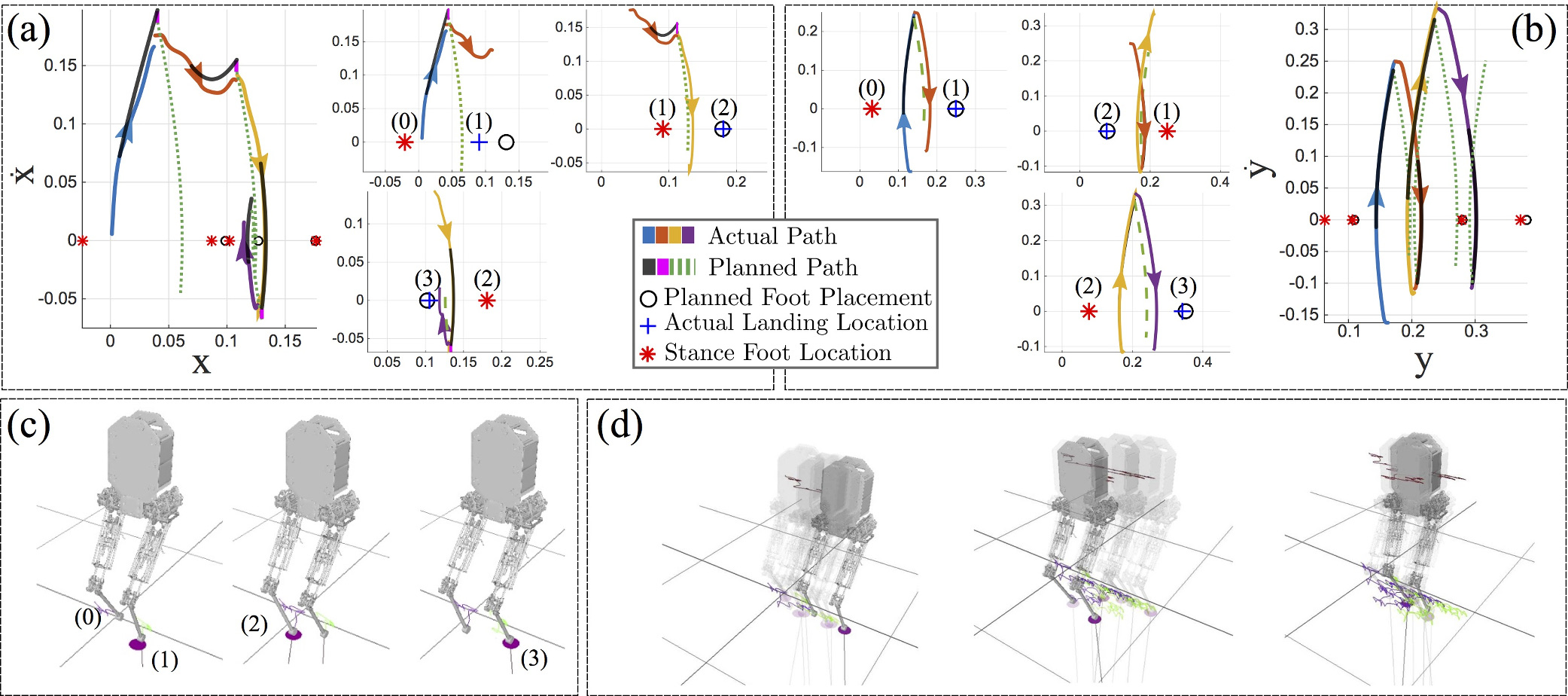}
\caption{{\bf 3D Stabilization in Simulation.} Subfig. (a) displays three steps of walking from forward COM phase space perspective, while (b) displays the lateral COM phase portrait. In both figures the smaller axes highlight the planned versus actual trajectory for each of the three steps. The steps are shown in (c) using the simulation graphics provided by the SrLib Multi-Body Dynamic Simulation Environment. Subfig.~(d) shows a longer time period of walking from several angles, and demonstrates the generally stationary behavior.}
\label{fig:planner_online}
\end{figure*}
As illustrated in Fig.~\ref{fig:phase_explain}, the planner begins calculating the landing location when 80\% of the lifting phase is reached in the robot state machine's progression. As such, the planner continuously re-plans in an online fashion to correct for trajectory deviations. Using the current estimate of the center of mass velocity and position, it numerically integrates forwards in time to predict its COM position and velocity when its stance foot and swing foot will switch roles. This time, position, and velocity is known as the switching state, \myfin\ in Fig.~\ref{fig:phase_explain}. Yet this is not the state from which the planner begins predicting the COM behavior for the upcoming swing. Instead, the planner applies a subtle modification to the switching state which represents the effect of landing. After applying this model, we arrive at the ``post-impact state'', \myimpact\ in Fig.~\ref{fig:phase_explain}. This point represents the planner's best guess at the center of mass position and velocity immediately after the leg switch. 

The implementation of the planner enforces the choice of a value for the reversing time, $t'$. This time value remains constant for every step, thus the user needs only to specify a single parameter for the planner to operate. As of now, $t'$ is manually chosen and as we show in the simulations it is able to stabilize the biped for an arbitrary long number of steps. The planner then finds and returns the footstep location which causes the robot's COM velocity to reach zero, $t'$ seconds after the foot switch, starting from the post-impact state. For each potential footstep location considered, the planner integrates forward in time starting from the post-impact state as suggested in Fig.~\ref{fig:phase_explain}f, returning the velocity after $t'$ seconds. This integration can be viewed as a function mapping footstep location to a future velocity, and it is this function over which we search for a zero crossing via bisection. Since we use bisection, the number of integrations actually performed is very low, however the process relies on the monotonicity of the relationship between footstep location and the velocity after $t'$ seconds of integration. If the height surface is planar, then this relationship is linear. 

\subsubsection{2D Velocity Reversal for 3D walking}
Choosing a footstep for 3D walking is, under certain circumstances, identical to choosing the footstep for 2D walking in two orthogonal directions. We take advantage of this interpretation to extend our constant time to velocity reversal planner to 3D, as we split 3D walking to a forward, $x$, phase space and a lateral, leftwards or $y$, phase space. While we could potentially allow different $t'$ parameters for the two planes, we use $t'=0.24$ seconds for our particular robot in both cases. This short time constant is indicative of the highly dynamic motion we are attempting, and is chosen to be as short as possible since faster stepping allows disturbances to be counteracted sooner.

Since an attempt to cross feet laterally could result in a leg-leg collision, we slightly modify the procedure in order to reduce the likelihood of such an event. The $y$ phase space and the $y$ component of the footstep location is always computed before that of $x$, so that the y phase space is given the opportunity to modify the time of swing to be used by the x phase space process. This is accomplished in the process of computing the switching state in the y phase plane. If the y velocity hits a maximum $|\dot y|$ limit of $0.65\ m/s$ during the integration, then the entire step is shortened such that the switching point occurs where the integration reaches the maximum y speed. If $|\dot y|$ at the default switching time is less than $0.1m/s$, the planner extends the step duration until the switching state has a $y$ speed of at least this value. When the $y$ phase plane process adjusts the switching time, the $x$ phase plane process uses this new switching time to find its switching state. Since the switching time represents the point at which the support leg changes, it cannot differ between the two directions. 

Although the $y$ directional impact does not appear to have a bias, an $x$ directional impact bias of $-0.01 m/s$ appears in the simulation results and is included in the planner, though it is smaller than that of the experiment.

Fig.~\ref{fig:planner_online} provides the results of our 3D walking algorithm in simulation, and demonstrates that the strategy stabilizes the velocity of the 3D simulation robot for an arbitrary long number of steps. Note that in step (1) of Fig.~\ref{fig:planner_online} the planner attempts to reverse velocity in $x$ and yet the velocity remains positive after the step has been made. This is due in part to the inaccurate landing of the footstep (1), which was not far enough forward to reverse the motion. This is a fairly common problem, especially at low speeds when the footsteps are very close together.

\section{Experimental Results and Assessment}
\label{sec:experiment}
\begin{figure*}
\centering
\includegraphics[width = 0.95\linewidth]{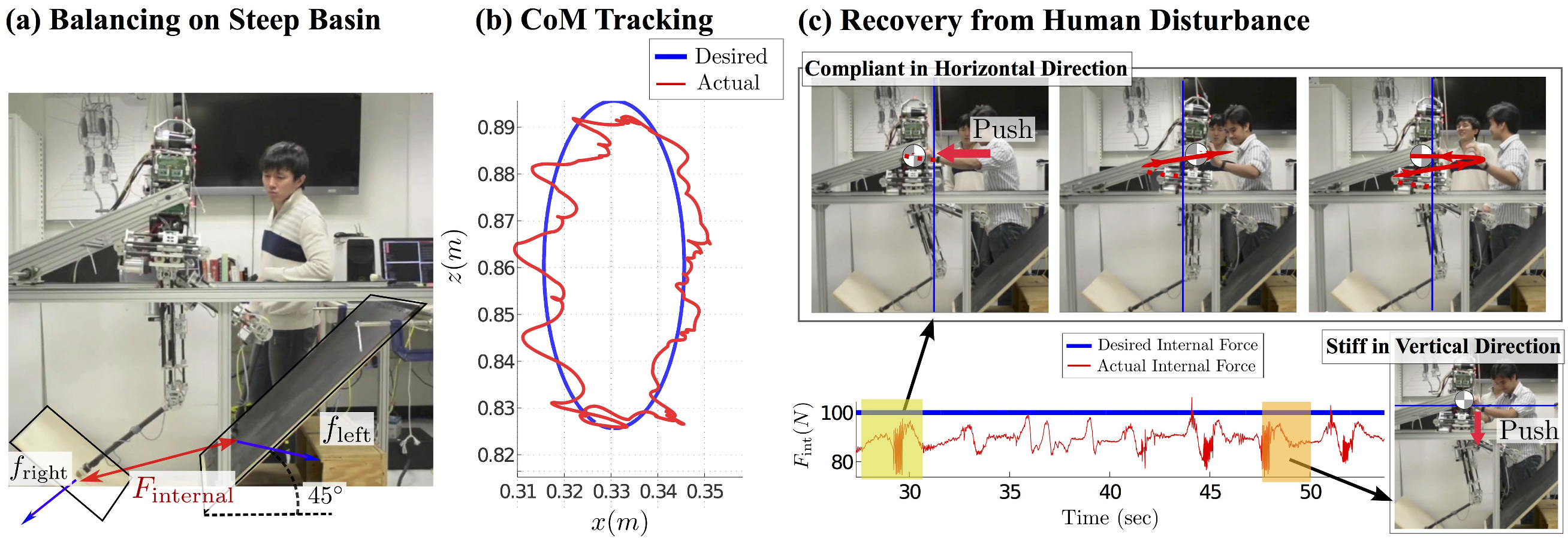}
\caption{{\bf Elliptical Trajectory Tracking and Human Disturbance Rejection on High Pitch Split Terrain:} Subfigure (a) shows Hume standing between two inclined wooden panels and tracking a position task with its center of mass. This position set-point follows a constant velocity trajectory along an elliptical path shown, along with the measured COM path, in subfigure (b). In Subfigure (c) the COM has a different impedance in the horizontal and vertical directions. When the robot is pushed backwards it moves as though the center of mass were connected to a low-spring-constant spring, whereas when the robot is pushed downwards it reacts as though connected to its set point by a far stiffer spring. Due to the feedback regulation of internal forces, the biped does not fall down when disturbed with large external forces.}
\label{fig:int_test}
\end{figure*}
We present here experimental results supporting the ideas proposed in previous sections. A planarizing linkage system, shown in Fig.~\ref{fig:infra}, constrains motion to the sagittal plane in all experiments. In the first experiment, Hume stands on two wedges inclined inwards as shown in Fig.~\ref{fig:int_test}(a). The first experiment shows balancing on a split terrain robustly handling human interactions. It demonstrates a successful implementation of WBOSC in a biped robot with elastic actuators, and validates the performance of internal force control. In the second experiment, Hume implements a stepping task with the planarizer's slider locked in place, allowing only vertical motion, approximately, and pitching of the body. We use the stepping test to validate the contact transition technique introduced in Sec.~\ref{sec:transitions}. The stepping test also tests the DSP-level gain scheduling strategy we used. In the final experiment, we show undirected dynamic walking with the proposed continuous re-planning method. As a final comment before describing further details, we note that in all controllers used in the experiments we omit the implementation of Coriolis/centrifugal terms represented by the symbols $\mu\task^*$ in Equation~(\ref{eq:task-space-dynamics}) and $\mu_i$ in Equation~(\ref{eq:internal-force-feedback}).


\subsection{Balance in a High Pitch Split Terrain}
In this experiment, the Hume biped balances on a high pitch terrain composed of two $45^{\circ}$ wedges angled in towards the robot to create a convex floor profile. As far as we know, this is the first time a biped robot has been reported under conditions that strictly require internal force to remain standing. The robot's tasks were to maintain a $100 N$ internal force pushing outwards against the two contact points, a desired impedance task for the center of mass point, and a desired impedance task for the body orientation. By controlling this internal force, Hume was able to avoid slipping while accurately adjusting its COM position. This experiment is divided into two sub-experiments: Hume was made to follow a time-varying COM trajectory which followed an elliptical path in the sagittal plane, as shown in Fig.~\ref{fig:int_test}(b); and Hume was made to hold a Cartesian impedance task on the COM which had low stiffness horizontally and high stiffness vertically as shown in Fig.~\ref{fig:int_test}(c). In the second sub-experiment the robot was perturbed by external disturbances in the form of human pushes. In this test, $x_{\rm task}^d = [{\rm COM}_x, \ {\rm COM}_z,\ q_{Ry},\ q_{Rx}]^T$, $F_{\rm task}^d = [0]_{4\times 1}$ and $F_{\rm int}^d = 100N$, using the notation of Fig.~\ref{fig:control}. Gains are summarized in Table~\ref{tb:internal_gain_set}.
\begin{table}
\centering
Position Gain
\begin{tabular}{>{\centering}m{0.17\columnwidth} %
                >{\centering}m{0.17\columnwidth} %
                >{\centering}m{0.17\columnwidth} %
                >{\centering}m{0.17\columnwidth} @{}m{0pt}@{}}
\specialrule{1.3pt}{1pt}{1pt}
& $K_x$ & $I_x$ & $D_x$ & \\ [2.0mm]
\hline
${\rm COM}_x$ & 15.0 & 0.0 & 0.0 & \\[1.5 mm]
${\rm COM}_z$ & 200.0 & 30.0 & 3.0 & \\[1.5 mm]
$q_{Ry}$ & 150.0 & 15.0 & 7.0 & \\[1.5 mm]
$q_{Rx}$ & 250.0 & 15.0 & 1.0 &\\[1.5 mm]
\hline
\vspace{3mm}
\end{tabular}
\\ Torque and Internal Force Gain
\begin{tabular}{>{\centering}m{0.242\columnwidth} %
                >{\centering}m{0.242\columnwidth} %
                >{\centering}m{0.242\columnwidth} @{}m{0pt}@{}}
\specialrule{1.3pt}{1pt}{1pt}
$K_{P, \tau}$ & $K_{I, tau}$ & $K_F$ & \\ [2.0mm]
\hline
50 & 0 & 1.0 & \\[1.5 mm]
\hline
\end{tabular}
\vspace{2mm}
\caption{{\bf Gain Set for the Internal Force Test.}  }
\label{tb:internal_gain_set}
\end{table}

\begin{figure}
\centering
\includegraphics[width = 8cm] {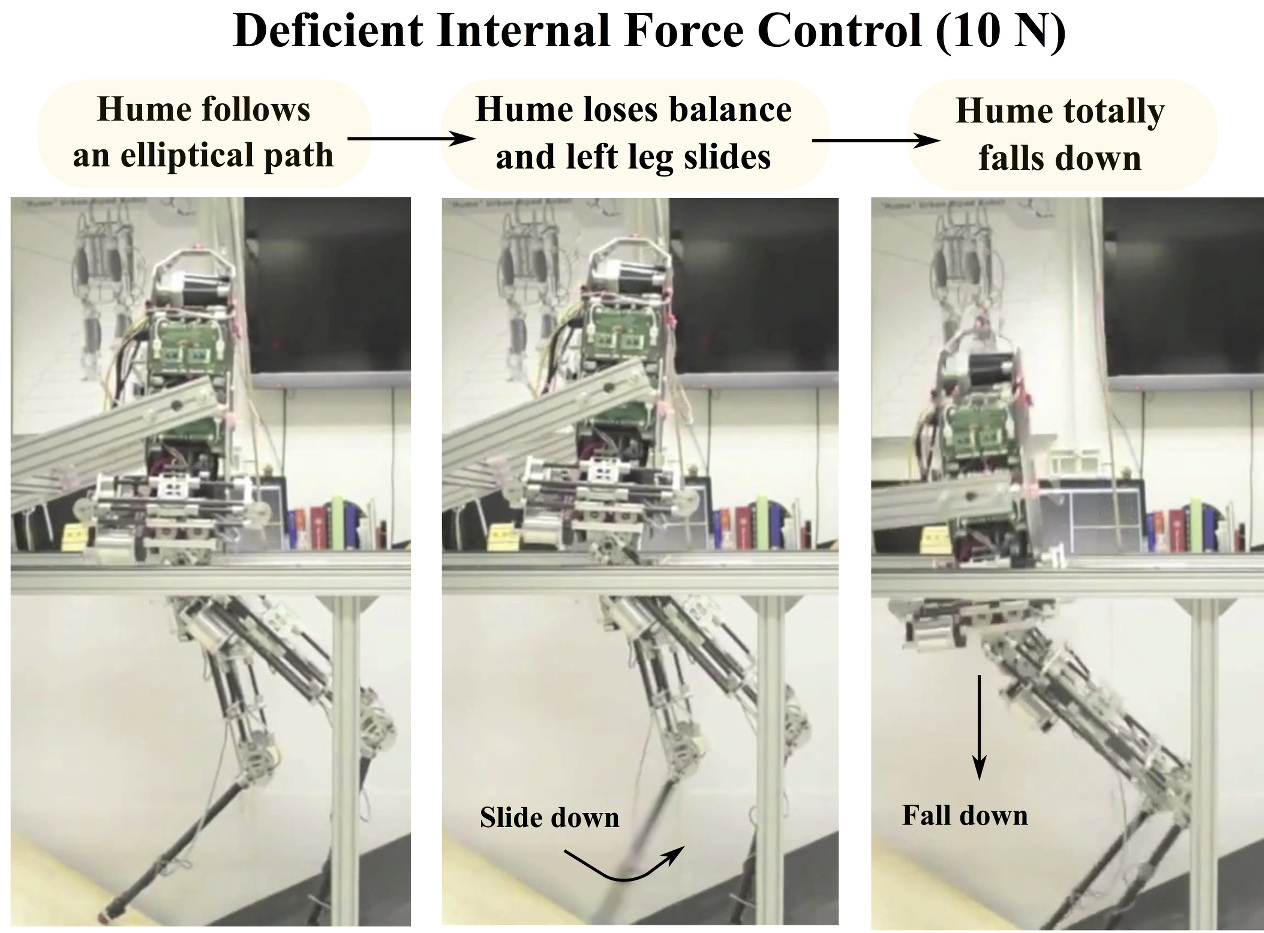}
\caption{{\bf Deficient Internal Force Control:} This figure demonstrates how Hume falls when the internal force is insufficient to maintain static friction at both feet. In this case the internal force was only 10 N which was not sufficient to overcome the effect of gravitational forces on the surfaces.}
\label{fig:lose_contact}
\end{figure}

Due to this feedback strategy, the errors between desired (black) and actual (red) internal forces are small enough to achieve a stable pose control. COM tracking performance is shown in Fig.~\ref{fig:int_test}(b). Although notable noise exists, the position error is bounded within $2 cm$. The second sub experiment shows that feedback control of internal forces allows them to persist despite external forces exerted by a human. As a testament to the compliance of our robot, the pushes we exert on the robot do not produce large deviations in the internal forces. 
Fig.~\ref{fig:int_test}(c) shows fast decay of disturbance transient phase, on the order of half a second. As shown in Fig.~\ref{fig:lose_contact}, without sufficient internal force the robot loses frictional contact with the wedges and this precipitates a fall.
These experiments ultimately show the effectiveness of closed loop internal force control as applied to maintaining a frictional contact. 

%

\begin{figure*}
\centering
\includegraphics[width = 0.95 \textwidth]{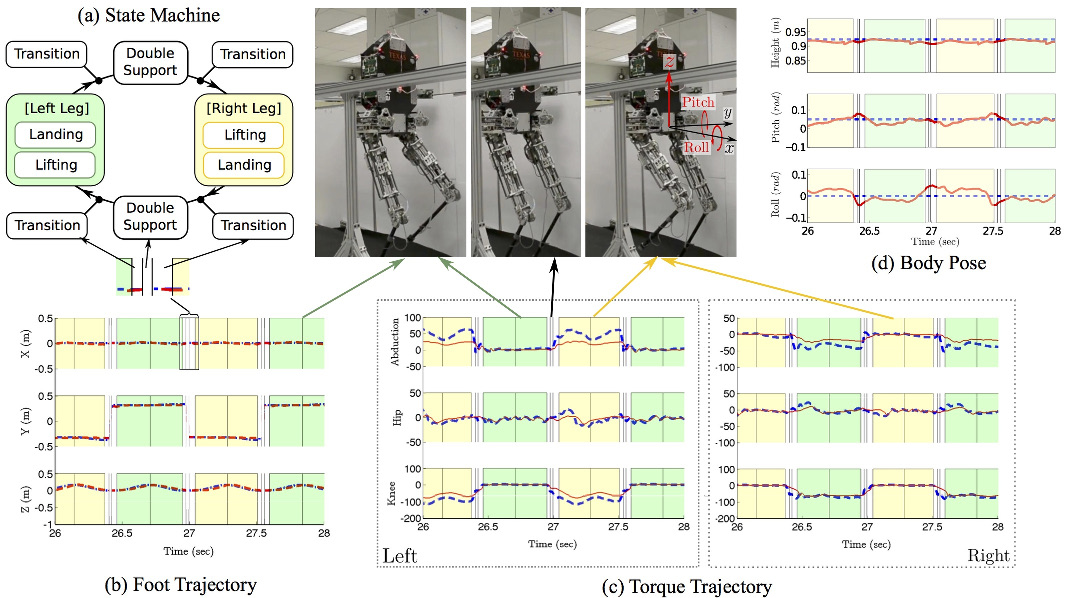}
\caption{{\bf State Machine and Stepping Motion Test:} (a) Motion within the stepping experiment is divided into three states: dual support, swing leg lift, and swing leg landing. Additionally, a transition state exists at the beginning of the lifting phase and the end of the landing phase in order to avoid the jerk caused by a sudden change of effective mass. Desired foot trajectories, blue, and sensor data, red, exhibit good tracking performance in Subfigure (b). In Subfigure (c), desired torque trajectories, blue, are plotted alongside sensor measured torques, in red. Background color indicates the phase of walking, with green corresponding to the left leg swing phase, yellow to the right leg swing phase, and white to the dual support phase. In Subfigure (d), position and orientation task tracking is plotted over a representative portion of the stepping experiment, with sensor data in red and desired values, dotted, in blue. Height refers to COM height.}
\label{fig:stepping_test}
\end{figure*}

\subsection{Stepping Test}

\subsubsection{Significance}
The stepping experiment offers a preliminary look at the performance of the control system with regard to achieving the zero dynamics of the pendulum model. It allows us to evaluate the performance of the transition strategy described in Sec.~\ref{sec:transitions}.

\subsubsection{Setup}
In this experimental setup, Hume's $x$ direction of motion was restricted by locking the sliding degree of freedom within the planarizing linkage. The robot's feet, placed roughly beneath the center of mass of the system, were lifted one at a time as though the robot were marching in place. Rather than using the planner, desired footstep location was held constant for each foot.

\subsubsection{WBOSC Inputs}
This stepping in place motion followed a time scripted state machine, shown in Fig.~\ref{fig:stepping_test}(a). Since the states are symmetric with respect to the supporting leg being either right or left, states are categorized in two different task sets with left and right single support having symmetric structures. The WBOSC task set differs between dual support and other phases of the stepping state machine. 


As shown in Fig.~\ref{fig:stepping_test}(d), the robot's tasks were to control COM height, body pitch and roll angles in dual support. Therefore, in dual support, $x_{\rm task}^d$ is $[{\rm COM}_z, \ q_{Ry}, \ q_{Rx}]^T$. In single support, we control foot position as well, therefore it changes to $[{\rm COM}_z, \ q_{Ry}, \ q_{Rx}, \ foot_x, \ foot_y, \ foot_z]^T$. Here, internal force feedback control is disabled due to the brief dual support, $0.02 s$. Transitions are used based on Sec.~\ref{sec:transitions}, with $F_{\rm task}^d=[0, \ 0, \ 0, w f_{r, x}, \ w f_{r, y}, \ w f_{r, z} ]^T$, where $w\in[0,1]$ is the scaling factor and $f_{r, -}$ is the expected reaction force in the dual support case.

\begin{figure*}
\includegraphics[width = 1.95\columnwidth]{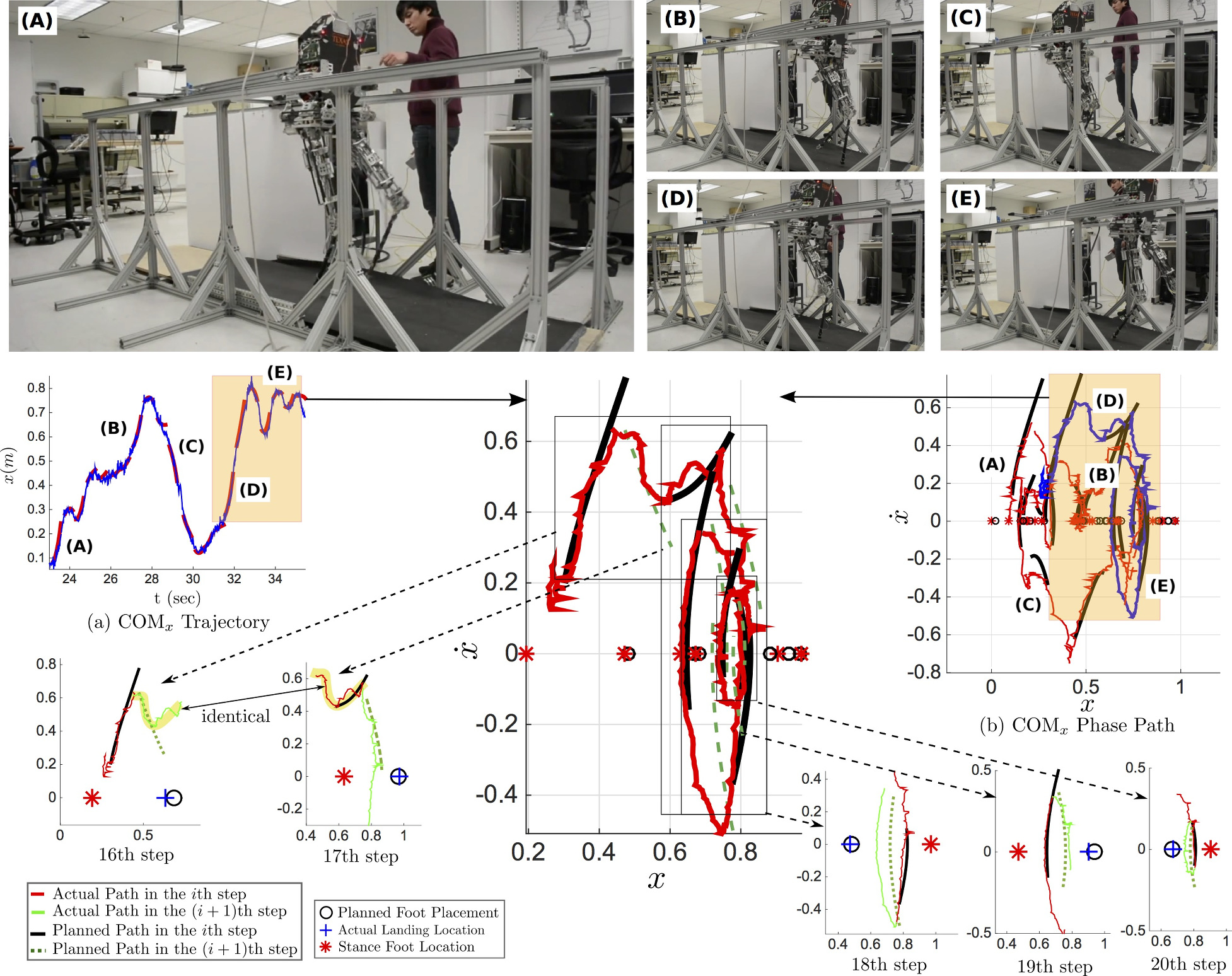}
\caption{{\bf Undirectional Walking.} Subfig.~(a) shows the $x$-directional trajectory of the COM. The blue line represents actual data while the red lines indicate the trajectories expected by the planner. Subfig.~(b) displays 23 steps of undirected walking. The central plot focuses on steps 16-21, of which 16-20 are expanded into individual step planning plots. In each step planning plot, a red line marks the actual COM path up to the switching state, and a green line continues the trajectory after the switch. The robot's initial stance foot in the step planning plot is denoted with a \mystance , the planned second footstep with a black circle, and the achieved second stance foot location with a blue cross. Therefore a green line in the $i$th step plot is the same path as the red line in the $(i+1)$th step plot. A black line and a dark green dotted line are, respectively, the PIPM model's predicted paths before and after the switching state. }
\label{fig:undirectional_walking}
\end{figure*}

\subsubsection{Data Analysis \& Conclusions}

The stepping task also used the gain scheduling algorithm described in Sec.~\ref{sec:gains}. With a positional COM error bounded within $2 cm$ and an orientation error within $0.15 rad$, as can be verified in Fig.~\ref{fig:stepping_test}(d), we can conclude that the controller has the potential to achieve closed loop standing through repeated stepping. Torque tracking performance for the low level controllers is displayed in  Fig.~\ref{fig:stepping_test}(b). The results show larger torque tracking errors in the single supporting leg than in the swing leg. When the gain scheduling algorithm enables the integral gain on torque tracking, during swing, this error becomes far smaller, which is necessary in order to overcome the friction internal to the actuator when moving the lightweight shank link of the swing leg.

Another notable result from Fig.~\ref{fig:stepping_test}(c) is that the torque transitions gently change between each step, as a result of the contact transitioning procedure. Even when the command changes from $100\ N$ to $0\ N$ this transition is handled gracefully. 
To conclude, this experiment shows that by using torque transitions and gain scheduling the control structure presented in this paper, Hume achieved a smooth and accurate stepping motion under WBOSC.
\begin{table}
\centering
Position Gain in Dual Support 
\begin{tabular}{>{\centering}m{0.17\columnwidth} %
                >{\centering}m{0.17\columnwidth} %
                >{\centering}m{0.17\columnwidth} %
                >{\centering}m{0.17\columnwidth} @{}m{0pt}@{}}
\specialrule{1.5pt}{1pt}{1pt}
& $K_x$ & $I_x$ & $D_x$ & \\ [2.0mm]
\hline
${\rm COM}_z$ & 450.0 & 55.0 & 5.0 & \\[1.5 mm]
$q_{Ry}$ & 400.0 & 55.0 & 5.0 & \\[1.5 mm]
$q_{Rx}$ & 80.0 & 10.0 & 5.0 &\\[1.5 mm]
\hline
\vspace{4mm}
\end{tabular}
\\ Position Gain in Single Support
\begin{tabular}{>{\centering}m{0.17\columnwidth} %
                >{\centering}m{0.17\columnwidth} %
                >{\centering}m{0.17\columnwidth} %
                >{\centering}m{0.17\columnwidth} @{}m{0pt}@{}}
\specialrule{1.5pt}{1pt}{1pt}
& $K_x$ & $I_x$ & $D_x$ & \\ [2.0mm]
\hline
${\rm COM}_z$ & 470.0 & 60.0 & 10.0 & \\[1.5 mm]
$q_{Ry}$ 	  & 300.0 & 60.0 & 5.0 & \\[1.5 mm]
$q_{Rx}$ 	  & 100.0 & 10.0 & 5.0 &\\[1.5 mm]
\hline
$foot_x$	 	  & 900.0 & 100.0 & 100.0 & \\[1.5 mm]
$foot_y$ 	  & 900.0 & 100.0 & 50.0 & \\[1.5 mm]
$foot_z$ 	  & 905.0 & 100.0 & 50.0 &\\[1.5 mm]
\hline
\vspace{4mm}
\end{tabular}
\\ Default Torque Gain
\begin{tabular}{>{\centering}m{0.242\columnwidth} %
                >{\centering}m{0.242\columnwidth} %
                >{\centering}m{0.242\columnwidth} @{}m{0pt}@{}}
\specialrule{1.5pt}{1pt}{1pt}
 &$K_{P, \tau}$ & $K_{I, \tau}$ & \\ [2.0mm]
\hline
Every Joint & 65 & 0 & \\[1.5 mm]
\hline
\vspace{4mm}
\end{tabular}
\\ Torque Gain of Swing Leg
\begin{tabular}{>{\centering}m{0.242\columnwidth} %
                >{\centering}m{0.242\columnwidth} %
                >{\centering}m{0.242\columnwidth} @{}m{0pt}@{}}
\specialrule{1.5pt}{1pt}{1pt}
 &$K_{P, \tau}$ & $K_{I, \tau}$ & \\ [2.0mm]
\hline
Hip right & 200 & 22 & \\[1.5 mm]
Hip left & 180 & 15 & \\[1.5 mm]
\hline
Knee right & 260 & 35 & \\[1.5 mm]
Knee left & 255 & 30 & \\[1.5 mm]
\hline
\end{tabular}
\vspace{2mm}
\caption{{\bf Gain Set for the Stepping Test.}  }
\label{tb:stepping_gain_set}
\end{table}
\newcommand{\fwt}{0.5\columnwidth}
\subsection{Undirected Walking with Online Re-planning Method}
%
%
The undirected walking experiment is designed to test the balancing ability of our system, testing both the continuous time feedback of our WBOSC implementation and the discrete time feedback from the footstep planner. In this experiment, Hume continually steps forward or backward in order to remain upright despite its inherently unstable dynamics. The abduction/adduction joint of the hip, unlike the other experiments, was fixed using joint level position control to simplify the problem. The experimental setup, as well as the data from this experiment, are shown in  Fig.~\ref{fig:undirectional_walking}.

Since the abduction/adduction joint is locked, $x_{\rm task}^d$ does not include roll motion control, and is thus $[{\rm COM}_z, \ q_{Ry}]^T$. In single support, it becomes $[{\rm COM}_z, \ q_{Ry}, \ foot_x, \ foot_z]^T$. Additionally, since the dual support period is only $0.079 s$ long, internal force feedback control is also disabled. Transitions are implemented identically to the way they were implemented in the stepping test, and as described in Sec.~\ref{sec:transitions}.

The difficulty of this challenge, when constrained to the sagittal plane, is highly dependent on the amount of time the robot spends in dual contact, which is naturally stable. We spend an almost trivial amount of time in dual support, limited primarily by the speed at which our contact transitions can proceed. The experimental velocity data used in Fig.~\ref{fig:undirectional_walking} are filtered by a steady state PIPM based observer. This filtered data are used only by the planner, and are not fed back by the WBOSC implementation.

\begin{table}
\centering
Position Gain in Dual Support 
\begin{tabular}{>{\centering}m{0.17\columnwidth} %
                >{\centering}m{0.17\columnwidth} %
                >{\centering}m{0.17\columnwidth} %
                >{\centering}m{0.17\columnwidth} @{}m{0pt}@{}}
\specialrule{1.5pt}{1pt}{1pt}
& $K_x$ & $I_x$ & $D_x$ & \\ [2.0mm]
\hline
${\rm COM}_z$ & 350.0 & 35.0 & 10.0 & \\[1.5 mm]
$q_{Ry}$ & 150.0 & 35.0 & 5.0 & \\[1.5 mm]
\hline
\vspace{4mm}
\end{tabular}
\\ Position Gain in Single Support
\begin{tabular}{>{\centering}m{0.17\columnwidth} %
                >{\centering}m{0.17\columnwidth} %
                >{\centering}m{0.17\columnwidth} %
                >{\centering}m{0.17\columnwidth} @{}m{0pt}@{}}
\specialrule{1.5pt}{1pt}{1pt}
& $K_x$ & $I_x$ & $D_x$ & \\ [2.0mm]
\hline
${\rm COM}_z$ & 350.0 & 35.0 & 10.0 & \\[1.5 mm]
$q_{Ry}$      & 150.0 & 35.0 & 10.0 & \\[1.5 mm]
\hline
$foot_x$	 & 300.0 & 50.0 & 15.0 & \\[1.5 mm]
$foot_z$ 	  & 300.0 & 50.0 & 10.0 &\\[1.5 mm]
\hline
\vspace{4mm}
\end{tabular}
\\ Default Torque Gain
\begin{tabular}{>{\centering}m{0.242\columnwidth} %
                >{\centering}m{0.242\columnwidth} %
                >{\centering}m{0.242\columnwidth} @{}m{0pt}@{}}
\specialrule{1.5pt}{1pt}{1pt}
 &$K_{P, \tau}$ & $K_{I, \tau}$ & \\ [2.0mm]
\hline
Both Hip Joints & 94 & 0 & \\[1.5 mm]
Both Knee Joints & 98 & 0 & \\ [1.5mm]
\hline
\vspace{4mm}
\end{tabular}
\\ Torque Gain of Swing Leg
\begin{tabular}{>{\centering}m{0.242\columnwidth} %
                >{\centering}m{0.242\columnwidth} %
                >{\centering}m{0.242\columnwidth} @{}m{0pt}@{}}
\specialrule{1.5pt}{1pt}{1pt}
 &$K_{P, \tau}$ & $K_{I, \tau}$ & \\ [2.0mm]
\hline
Hip right & 200 & 22 & \\[1.5 mm]
Hip left & 200 & 22 & \\[1.5 mm]
\hline
Knee right & 275 & 35 & \\[1.5 mm]
Knee left & 270 & 32 & \\[1.5 mm]
\hline
\end{tabular}
\vspace{2mm}
\caption{{\bf Gain Set for the Undirected Walking Test.}  }
\label{tb:walking_gain_set}
\end{table}

\begin{table}
\centering
\begin{tabular}{>{\centering}m{0.4\columnwidth} %
                >{\centering}m{0.27\columnwidth} @{}m{0pt}@{}}
\specialrule{1.5pt}{1pt}{1pt}
Phase & Time (sec) & \\ [2.0mm]
\hline
 Transition & 0.02 & \\[1.5 mm]
 Lifting & 0.23 &  \\[1.5 mm]
 Landing & 0.26 &  \\[1.5 mm]
 Double Support & 0.079 & \\ [1.5 mm]
\hline
$t'$ & 0.25 & \\[1.5mm]
\hline
\end{tabular}
\vspace{2mm}
\caption{{\bf Time Parameters for the Undirected Walking Test}  }
\label{tb:time_undirectional_walking}
\end{table}
%
A critical test of our planning methodology is whether or not the simplified model we used to approximate the zero dynamics of the controlled system are an accurate predictor of the future center of mass state. And demonstrated by Fig.~\ref{fig:undirectional_walking}, the model predicts the continued evolution of the current step accurately, and the evolution of the second step with a much larger propensity to deviate widely. This is because the landing event is a major source of uncertainty. Not only is the controller performing a transition maneuver at this point, but the landing error of the foot is not insignificant relative to the length of an average step in both space and time. 

Noticing that the robot reacts to contact transitions with a consistent bias, we model a $-0.1\ \rm m/s$ jump in velocity at each transition to reflect this reality. This impact model separates the switching state from the post-impact state in Fig.~\ref{fig:phase_explain}. The five miniature phase plots in Fig~\ref{fig:undirectional_walking} verify that this simple methodology successfully predicts the ${\rm COM}_x$ path in the next step.

In the 16th step, Hume's foot deviates the planned location by $5.5 \ \mathrm{cm}$, and the planned future trajectory differs significantly from the truth, progressing beyond the foot location rather than changing direction in constant time as was the intention of the planner. This highlights the fact that the dynamics of the robot are highly sensitive to footstep position error, and also demonstrates that while the robot's footstep accuracy was very high in the stepping experiment, this accuracy decreases as the foot is required to follow faster landing trajectories with more significant forward motion. However, this mis-step is brought back under control in the next step and the COM slows down considerably by the 20th step. Once the robot has slowed down to this degree, it becomes nearly impossible to predict the direction it will fall, which causes the robot to drift sideways until it hits the mechanical travel limitation of the planarizer.


\section{Conclusion}


Operational Space Control was originally conceived for mobile manipulators to simultaneously control forces and accelerations in a dynamically consistent manner. Whole-Body Operational Space Control further extended this methodology for floating base humanoid robots with natural constraints. In particular, it defines the space of internal forces as a fully controllable task system that is orthogonal to the robot's acceleration manifold. This interpretation allows to employ feedback control techniques to regulate or track internal force reference inputs with good precision. For instance, in this paper we implement a proportional closed loop feedback control process to regulate sensor-based internal forces to a desired value for balancing on high pitch split surfaces. 
This case scenario exemplifies the need for a unified force/motion feedback control approach for this peculiar type of balancing which is facilitated by WBOSC. 

Point foot robots like ours cannot control their center of mass in single support without exploiting the state-dependence of the gravity and Coriolis vectors. Therefore it is difficult for them to implement locomotion strategies based on center of mass tracking such us those using the Zero Moment Point or the Capture Point. When we started this research we were motivated to attain non-periodic gaits. We were driven by applications such as rough terrain walking, jumping between vertical walls, and push recovery. At that time there were no algorithms suited for those type of behaviors in point foot robots. We devised a new rule-based algorithm, which we dubbed Phase-Space Planning~\cite{Sentis:11(ISRR)}, in which foot positions and apex velocities were known a priory. Phase space techniques were then employed to find transitions states between the steps. For this new work, we decided to extend phase space techniques to the general case of continuously stepping such that the robot's center of mass velocity could be quickly reversed. The overall effect is an undirected walk that stabilizes the robot through continuous replanning capabilities. Although such walking does not necessarily stay in place, our experiment has two goals: to create an environment for future push recovery, and to create an algorithm for 3D untethered balancing. 
This rule was effective in stabilizing the robot in a 3D simulation as well as in the physical system with a 2D planarizing constraint.

What are the advantages of calculating output joint torques on the SEAs versus, for instance, incorporating six axis force/torque sensors on the biped's feet? The main advantage is that we can implement sensor-based feedback control behaviors between multiple contacts on any part of the robot's body. For instance, a biped robot like ours could balance using its knees against surfaces instead of its feet. In that case, internal forces between the knees can be readily calculated using the joint torque sensors. This capability will enable bipedal robots to achieve remarkable levels of agility using any part of their bodies.


One of the main advantages of using WBOSC versus other controllers is that internal forces can be tracked with high fidelity based on sensor feedback. WBOSC is the first controller to have sensor-based feedback control capabilities on the internal forces. However, internal forces need to be computed such that the robot can balance on the steep terrain. We have chosen a high outward tension between the feet in contact such that they overcome gravity forces in our steep terrain test. However, more sophisticated techniques are recommended to automate the balancing process. In particular, the multicontact/grasp model presented in~\cite{Sentis:10(TRO)} provides a potential tool to extract internal forces given the desired center of mass behavior of the robot. More specifically, given desired center of mass behavior, we could use that model to extract reaction forces that fulfill friction constraints and then project the resulting reaction forces into the internal force manifold for feedback control. 

While our experimental results in this paper have focused on the robot under planarizing constraint, it is an aim of our group to recreate the 3D balancing behavior of the simulation in physical hardware. This may require significant performance improvements in the sensing and state estimation system, and represents a more difficult experimental procedure as well. It may also become necessary to upgrade the abduction motors, two of which burned in the process of tuning the robot for planer walking. 



In summary, we have presented a method to place the internal forces of multi-contact under feedback regulation and validated that this method can prevent slipping in steeply angular terrains. We have verified the practicality of WBOSC for maintaining a zero dynamic manifold in an under-actuated system which permits discrete time footstep control. We have developed a discrete time controller which can stabilize point foot robots in 3D simulation, and we have shown that it can stabilize our physical robot in 2D.

\appendices

\section{Basics of Whole Body Operational Space Control} 
\label{sec:appendice}
Whole-Body Operational Space Control was first laid out in \cite{sentis2007synthesis} with a further exploration of internal forces to be published later in \cite{Sentis:10(TRO)}. Interested readers should refer to these sources for a description of the theoretical background complete with proofs for the concepts below, as space limits us to a cursory overview of WBOSC as applied to bipedal robots.
\newcommand{\njoints}{{n_{\mathrm{joints}}}}
Modeling biped poses entails representing not only the state of each joint 
$q_1,\dots,q_\njoints$, but also the states of the robot as an object in space. 
We choose to parameterize this as a $6$-dimensional floating joint between the world and the robot's hip coordinate system with state vector $q_b\in\mathbb{R}^6$. 
Combining the robot and base states into a single vector we arrive at 
$q\in \mathbb{R}^6\oplus\mathbb{R}^{\njoints}=\mathbb{R}^{\ndofs}$,
the generalized joint vector.
The joint torques can only directly effect the joints themselves, and not the floating base dynamics, so we say an under-actuation matrix 
$U \in \mathbb{R}^{\big(n_{\rm dofs}-6\big)\times n_{\rm dofs}}$  
selects from the joint vector to the subspace of actuated joints.
\begin{equation}
q\act = U \; q,
\end{equation}
with $q\act \in \mathbb{R}^{\njoints}$ being the robot's actuated joints. The dynamics of the robot's generalized joints can be described by a single, linear, second order differential equation
\begin{equation}
A\ddot q + b + g = U^T \tau_{\rm control}
\label{eq:unconstrained_dynamics}
\end{equation} 
where $\{A,b,g\}$ represent the inertia matrix, Coriolis/centrifugal forces, and gravitational forces, respectively while $\tau_{\rm control}$ is the desired control command applied to the output joints of the robot. Without considering contacts, or the subtle non-holonomic effect of angular momentum, joint torques would have no effect on the geometrically uncontrollable six dimensional subspace of the generalized joints: 
$ \{ z\in\mathbb{R}^{\ndofs}:z^TA^{-1}U^T=0\} $.
However we can sometimes gain the ability to control more of this space due to contact constraints. 

We consider two contact cases for point-foot biped robots: single support in which one foot is in contact, and dual support where the robot is supported by both feet. In the single support phase we describe the contact via a support Jacobian $J_s \in \mathbb{R}^{3\times n_{\rm dofs}}$, which maps from generalized joint velocity to the velocity of the constrained foot point in Cartesian space. When considering dual contact, our support Jacobian represents twice as many constraints. Since this generalized point, either the single foot point in $\mathbb{R}^3$ or the dual foot point in $\mathbb{R}^6$, is constrained, we know its acceleration must be zero. Substituting the constraint 
$J_s \ddot{q} + \dot J_s \dot q = \ddot x_{\rm foot(or\;feet)} = 0$
and adding the associated co-state of constraint space reaction forces $\lambda$, the dynamics become
\begin{equation}\label{eq:lambda-dyn}
A\ddot q + b + g + J_s^T \lambda = U^T \tau_{\rm control},
\end{equation}
and we can find $\ddot{q}$ and $\lambda$ by solving the matrix equation
\begin{equation}
\begin{pmatrix}
A & J_s^T \\ J_s & 0
\end{pmatrix}
\begin{pmatrix}
\ddot{q} \\ \lambda
\end{pmatrix}
=
\begin{pmatrix}
U^T \tau_{\rm control} -b-g \\ -\dot J_s \dot q
\end{pmatrix}.
\end{equation}
Converting to upper diagonal form via Gaussian elimination we find
\begin{equation} \label{eq:constraintforces}
\begin{pmatrix}
A & J_s^T \\ 0 & I
\end{pmatrix}
\begin{pmatrix}
\ddot{q} \\ \lambda
\end{pmatrix}
=
\begin{pmatrix}
U^T \tau_{\rm control} -b -g \\[2mm] 
\overline J_s^{\,T} \left[U^T \tau_{\rm control} -b -g\right] +\Lambda_s \dot J_s \dot q\\[1mm]
\end{pmatrix},
\end{equation}
with $\Lambda_s \triangleq [J_sA^{-1}J_s^T]^{-1}$ and $\overline J_s \triangleq A^{-1} J_s^T \Lambda_s$. Substituting $N_s^T\triangleq I-J_s^T\Lambda_s J_s A^{-1}$ to more conveniently express the resulting constrained dynamic equation,
\begin{equation}\label{eq:constrained-dyn}
A \, \ddot q + N_s^T \left( b + g \right)+J_s^T\Lambda_s \dot J_s \dot q= \left(U N_s \right)^T \tau_{\rm control}.
\end{equation}
This can be viewed as constraining the dynamics to the dynamically consistent null-space of the constraint by defining the dynamically consistent pseudo inversion operator 
\begin{equation}
\overline{X} \triangleq A^{-1}X^T[X A^{-1} X^T]^{-1}
\end{equation} and observing that \begin{equation}
N_s = I - \overline J_s J_s
\end{equation}
is the null space projector of $J_s$ under dynamically consistent inversion such that $J_s N_s = 0$, $N_s A^{-1} J_s^T = A^{-1} N_s^T J_s^T = 0$, and $N_s N_s = N_s$.  

The above dynamic equation is only valid if we assume that the robot actuators are completely rigid and that no friction occurs in the joints. Obviously this is not the case for our series-elastic biped. However, the joint level torque controllers in our robot are designed to make the actuators behave close to an ideal torque source. In that case, the model above more closely approximates the real dynamics. 

There are also some subtleties associated with the inertia matrix in this model given imperfect series elastic actuators. In the standard model of whole body control the matrix would reflect only the inertia of the rigid bodies of the robot. Yet in the series elastic case this is equivalent to assuming ideal performance: torque sensor measurements perfectly following the desired torque. With the lower force gains we used in the experiments this is less reflective of reality, and we include the reflected rotor inertia of the motor rotors in the mass matrix. This addition better reflects the state of setting the motor current without force feedback, but in reality the system is somewhere in the middle of these two extremes. Fortunately the behavior of the robot is not especially sensitive to these slight variations on the mass matrix in practice. 

With those assumptions, Whole-Body Operational Space Control for an operational task representation, $p\task$, is defined by the differential kinematic equation
\begin{equation}
\dot p\task = J\task^* \; \dot q\act,
\end{equation}
where $J\task^* \triangleq J\task \overline{UN}_s \in \mathbb{R}^{n\task\times n\act}$ is the contact consistent task Jacobian and $J\task \in \mathbb{R}^{n\task\times \ndofs}$ is the unconstrained task Jacobian. The basic control structure for the single support phase of the biped is thus
\begin{equation}\label{eq:task-control}
\tau_{\rm control} = J\task^{*\,T} F\task,
\end{equation}
with $F\task$ being the entry point for feedback control laws to govern trajectories, applied forces, or combinations of the two in the operational space. For instance, when controlling an operational space position trajectory, we use the model-based control law
\begin{equation}\label{eq:task-space-dynamics}
F\task = \Lambda\task^* u\task + \mu\task^* + p\task^*,
\end{equation}
with $u\task$ being a desired acceleration for the operational reference. $\{\Lambda\task^*,\mu\task^*,p\task^*\}$ are inertia, velocity-based forces, and gravity based forces in the operational space that can be found in the previous references. Under idealized perfect conditions, the effect of the above control command is the linearized closed loop dynamics,
\begin{equation}
\ddot x\task = u\task.
\end{equation}

In the case of double support, there appear closed loop effects between the legs of the robot in contact with the environment. Our previous work has thoroughly addressed this problem by creating structures to control the internal forces. Internal forces are defined as those that are orthogonal to joint accelerations. As such, internal forces do not produce any movement and only contribute to force generation within the closed-loop formed by the contacts. Analyzing the right hand side of Equation~(\ref{eq:constrained-dyn}), those forces correspond to the manifold
\begin{equation}
\big( U \, N_s \big)^T \; \tau_{\rm control} = 0,
\end{equation}
which reflect the cancellation of acceleration effects. Therefore, the torques that fulfill the above constraint belong to the null space of $(U\, N_s)$, which is defined by the projection
\begin{equation}
L^* \triangleq \Big(I-U\, N_s \, \overline{U\,N_s}).
\end{equation}
The torques associated with internal forces are those that do not contribute to net movement, i.e.,
\begin{equation}\label{eq:int-control}
\tau_{\rm control} = L^{*\,T} \tau_{\rm int},
\end{equation}
where $\tau_{\rm int}$ is the set-point for the internal forces. Thus, when simultaneously controlling operational space tasks and internal forces we superimpose the orthogonal structures of Equation~(\ref{eq:task-control}) and (\ref{eq:int-control}) yielding the whole-body operational space control command
\begin{equation}
\tau_{\rm control} = J\task^{*\,T} F\task + L^{*\,T} \tau_{\rm int}. 
\end{equation}

Internal forces can be physically defined as linear forces and mutually canceling reaction moments between pairs of supporting contacts. As explained in our previous works, such forces are expressed via the equation,
\begin{equation}\label{eq:wbosc:mapping_internal_reaction}
F_{\rm int} = W_{\rm int} \; F_r,
\end{equation}
where $F_r$ is the set of all reaction forces on the environment and $W_{\rm int}$ is a matrix containing geometric transformations and selection criteria to extract the internal forces. Using this mapping, we demonstrated that the dynamics of internal forces correspond to
\begin{equation}\label{eq:wbosc:internal_forces}
F_{\rm int} \, = \, \Big(\overline{J}_{i|l}^{\,*}\Big)^T \, \Gamma_{\rm int} + F_{{\rm int},\{t\}} - \mu_i - p_i,
\end{equation}
where $\overline{J}_{i|l}^{\,*} \triangleq \left( L^* U \overline J_s W_{\rm int}^T \right)$. Additionally, $F_{{\rm int},\{t\}}$ are internal forces induced by task behavior, i.e. $F_{{\rm int},\{t\}} \triangleq W_{\rm int} \overline J_s^T J\task^{*\,T} F\task$, and $\mu_i$ and $p_i$ are Coriolis/centrifugal and gravitational effects on the internal forces. Inverting the above equation, we derive the torques needed to accomplish a desired internal force,
\begin{equation}\label{fig:int-for}
\Gamma_{\rm int} \, = \; J^{\,*\,T}_{i|l} \Big(F_{\rm int, ref} -  F_{{\rm int},\{t\}} + \, \mu_i + \, p_i\Big),
\end{equation}
where $J_{i|l}^*$ is the Moore–Penrose left-pseudoinverse of $\overline{J}_{i|l}^{\,*}$, also referred to as the reduced Jacobian of internal forces acting on the contact closed loops, and $F_{\rm int, ref}$ is the vector of desired internal forces which we use as an entry point to control internal forces.

\bibliographystyle{IEEEtran}
\bibliography{../tro.bib}

\end{document}